\documentclass{article}
\usepackage[square, numbers]{natbib}
\usepackage[preprint]{tmlr}

\usepackage{amsmath,amsfonts,bm}

\def\eqref#1{equation~\ref{#1}}

\def\1{\bm{1}}

\DeclareMathAlphabet{\mathsfit}{\encodingdefault}{\sfdefault}{m}{sl}
\SetMathAlphabet{\mathsfit}{bold}{\encodingdefault}{\sfdefault}{bx}{n}

\usepackage[utf8]{inputenc} 
\usepackage[T1]{fontenc}    
\usepackage{float}
\usepackage{hyperref}       
\hypersetup{hidelinks}
\usepackage{url}            
\usepackage{booktabs}       
\usepackage{longtable}
\usepackage{amsmath}
\usepackage{amsfonts}       
\usepackage{amssymb}
\usepackage[nounderscore]{syntax}
\usepackage{nicefrac}       
\usepackage{microtype}      
\usepackage{xcolor}         
\usepackage{graphicx}
\definecolor{LightGray}{RGB}{245, 245, 245}
\usepackage[whole]{bxcjkjatype}

\title{KamonBench: A Grammar-Based Dataset for Evaluating Compositional
  Factor Recovery in Vision-Language Models}

\author{%
  Richard Sproat, Stefano Peluchetti\\
  Sakana.ai \\
  \texttt{\{rws,stepelu\}@sakana.ai}
}

\begin{document}

\maketitle

\begin{abstract}
        Kamon (家紋, family crests) are an important part of Japanese culture and a
        natural test case for compositional visual recognition: each crest combines a
        small number of symbolic choices, but the space of possible descriptions is
        sparse. We introduce KamonBench, a grammar-based image-to-structure
        benchmark with 20,000 synthetic composite crests and auxiliary component
        examples. Each composite crest is paired with a formal kamon description
        language---\emph{kamon yōgo} (家紋用語)---description, a segmented
        Japanese analysis, an English translation, and a
        non-linguistic program code. Because each synthetic crest is generated
        from known factors, namely container, modifier, and motif, KamonBench
        supports evaluation beyond caption-level accuracy: direct program-code
        factor metrics, controlled factor recombination splits, counterfactual
        motif-sensitivity groups under fixed container-modifier contexts, and
        linear probes of factor accessibility. We include
        baseline results for a ViT encoder with a Transformer decoder, two VGG
        n-gram decoders with and without learned positional masks, and fine-tuned
        Qwen3-VL-4B and Qwen3-VL-8B models. KamonBench
        therefore provides a controlled testbed for sparse compositional visual
        recognition and factor recovery in vision-language models.
\end{abstract}

\section{Introduction}
\label{sec:intro}

\emph{Kamon} (家紋, `family crests') have been a part of Japanese culture since
at least the 13th century Kamakura period
\citep{Stroehl:06,Dower:71,Chikano:93,
  Morimoto:06,Takasawa:08,Morimoto:13,Phillips:18,Takasawa:21,Sproat:23}.  The
original use was probably as an easily identifiable mark for personal property
\citep{Takasawa:21},
but soon, samurai began using family crests to distinguish clans
 during battles. In this usage in particular, kamon were functionally the same
 as coats of arms in Europe.
Traditionally European heraldry has been associated with nobility, and
to a large extent that remains true today.  For example, in England, if you
want to register a coat of arms, you must apply to the \emph{College of
Arms} and prove that you are \emph{armigerous}, i.e. have the right to bear
arms \citep{Fox-Davies:09,Friar:Ferguson:93,Slater:02}. In contrast, while kamon
were originally associated with noble and warrior families, they have become
democratized, so that today almost every family has its own family crest. Kamon
are a common sight in Japanese cemeteries, where crests adorn practically every
tomb.

\begin{figure}[t]
\centering
\begin{tabular}{c|ccc}
\includegraphics[height=2cm]{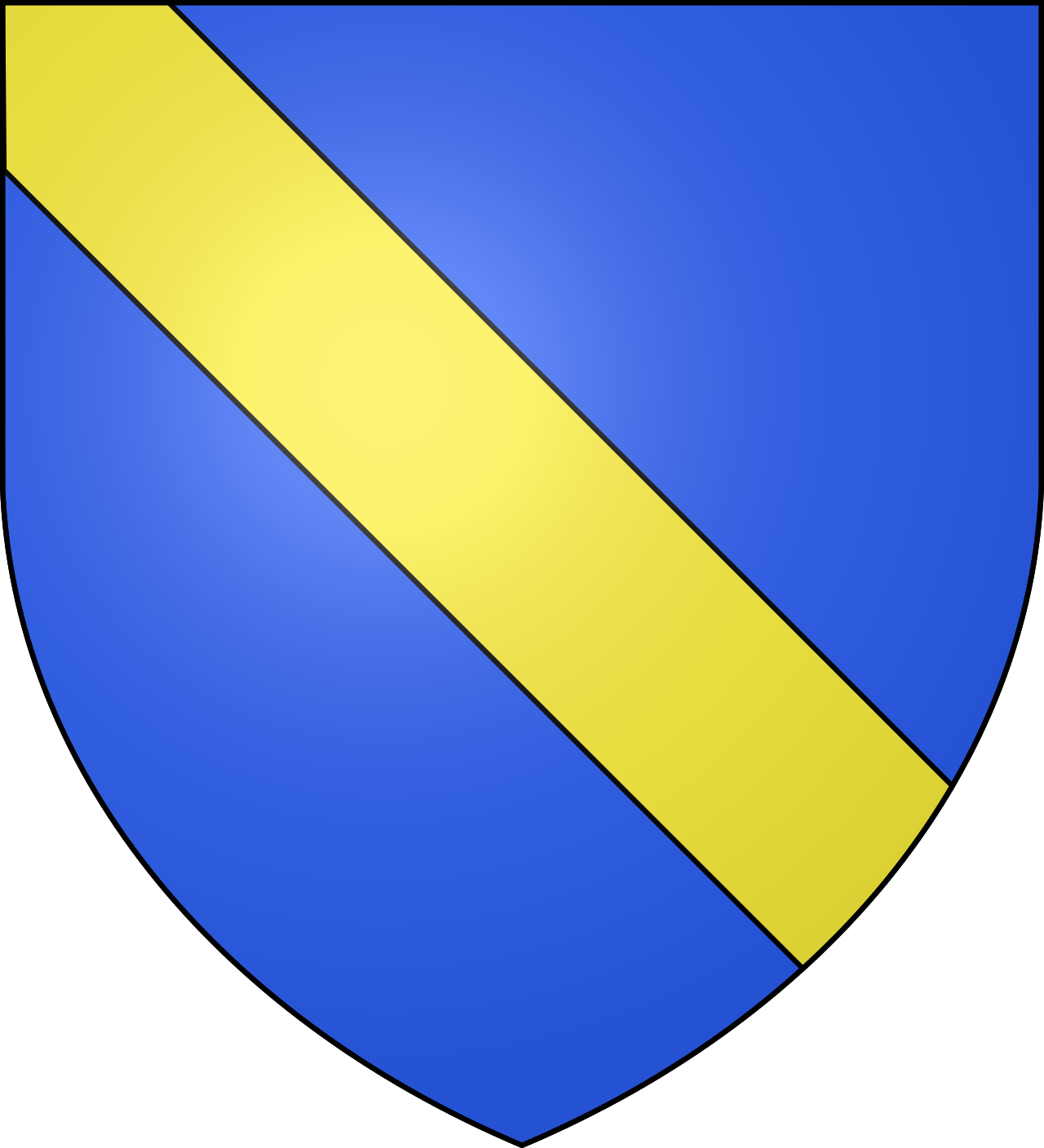}
 &
a) \includegraphics[width=2cm]{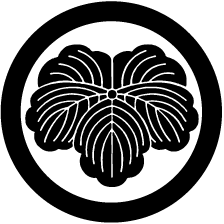} &
b) \includegraphics[width=2cm]{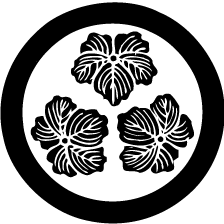} &
c) \includegraphics[width=2cm]{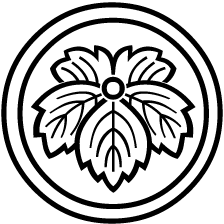} \\
 &
d) \includegraphics[width=2cm]{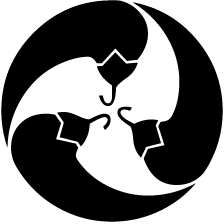} &
e) \includegraphics[width=2cm]{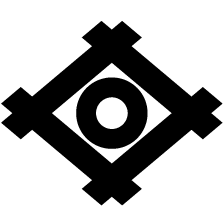} &
f) \includegraphics[width=2cm]{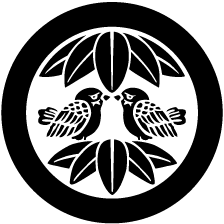} \\
\end{tabular}
\caption{Left: Example of British heraldry, with a simple shield described in
  blazon as \emph{azure a bend or}.  Source: Bear17
  ({\scriptsize\url{https://commons.wikimedia.org/wiki/File:Azure,_a_bend_Or.svg}}),
  CC~BY-SA~3.0.\\
  Right: Real kamon designs (not from the synthetic KamonBench dataset; shown
  to illustrate the broader vocabulary that motivates the benchmark):
  a) circle with ivy (丸に蔦);
  b) circle with three bottoms-together ivy (丸に尻合わせ三つ蔦);
  c) circle with demon ivy (総陰丸に鬼蔦);
  d) chili pepper swirl (唐辛子巴);
  e) well frame with snake eyes (井桁に蛇の目);
  f) circle with two rows of five bamboo leaves and facing sparrows (丸に二弾五
  枚笹に対い雀). Source:
  \cite{Sproat:26}.}
\label{fig:examples}
\end{figure}

Both European heraldry and kamon are associated with \emph{formal languages}
that are used to describe the arrangement of motifs within a coat of arms or
crest. In British heraldry the formal language is called \emph{blazon} and
consists of a rigidly defined set of terms for tinctures (which are part of the
graphical vocabulary of heraldry), motifs and their arrangements. For example,
Figure~\ref{fig:examples}, left panel, shows a simple coat of arms described by
the blazon \emph{azure, a bend or}. Here \emph{azure} means `blue', \emph{bend}
denotes the diagonal stripe on the shield, and \emph{or} means `gold'---much of the vocabulary of British heraldry deriving from French.  The
corresponding formal language for kamon is called \emph{kamon y\=ogo} (家紋用
語), which we will henceforth refer to as `kamon description language' (KDL).
KDL is less tightly constrained than blazon, but nonetheless is restricted in
the ways one can refer to motifs and their arrangements, and the ways in which
the descriptions are constructed.

Like European heraldry, kamon have hundreds of motifs, and these motifs can be
combined in various ways.  In many family crests, one or more motifs are
contained within an outer shell such as some type of ring, or a
polygonal figure such as a square or hexagon. We call this outer shell a
\emph{container}. Figure~\ref{fig:examples}, right panel, shows various
examples of complex kamon along with their corresponding KDL.

Two other operation types are important.  A \emph{spatial arrangement} changes
how copies of a motif are positioned, for example by stacking three copies or
placing their heads or bottoms toward the center.  A \emph{modification}
changes the form or scale of a motif. For example, the `demon' (鬼)
modification, which is generally applied to plant motifs, means that the leaf
or flower of the plant is depicted with sharpened points \citep{Takasawa:21}.
See Figure~\ref{fig:examples}c, and compare the ivy motif in that image with the
basic ivy in Figure~\ref{fig:examples}a. Another modification, `bean' (豆),
means that the motif is reduced in size (see below,
Figure~\ref{fig:synth_examples}b). While there are hundreds of motifs for family
crests, the possibilities for spatial arrangements and modifications are
limited. Therefore, family crests and their analysis are highly constrained
problems. In what follows we use \emph{modifier} as the umbrella term for either
a spatial arrangement or a modification. A kamon crest is therefore characterized by
three factors: \emph{container}, \emph{modifier}, and \emph{motif}. When an
analysis distinguishes the two modifier subtypes,
we refer to spatial arrangements and modifications separately.

We introduce KamonBench, a grammar-based benchmark for evaluating factor
recovery and compositional generalization in vision-language models. We also
compare human performance with that of multimodal language models after
limited instruction.

\section{Kamon as a machine learning problem}
\label{sec:problems}

Kamon are an important part of Japan's cultural heritage, but
what makes them particularly interesting as a machine learning problem, and in
particular what makes them interesting as an image-to-structure problem?
Like standard image-to-text (or text-to-image) problems, kamon are complex in
that `scenes' may consist of multiple elements in various spatial arrangements,
and elements may themselves be modified in various ways. But unlike standard
image-to-text cases, the modifiers (in the umbrella sense introduced in
Section~\ref{sec:intro}) are relatively constrained. This translates, on the
language side, to there being a relatively large set of terms
corresponding to the set of basic motifs, and an additional set of a few dozen
terms to express modifiers of motifs.  Containers, motifs that may contain
other motifs, are also relatively constrained, again limited to a few dozen
cases. But since containment may be recursive, the set of kamon is
theoretically unbounded. Even without recursion, the present factor inventory
yields approximately 770,000 non-recursive combinations. Thus, kamon comprise a
large set of possible images, each related to a string in KDL.

At the same time, kamon represent a \emph{sparse data} scenario, since while
there are in principle many examples of image-text combinations found on various
sites on the internet, these are largely limited to the more common
crests, typically those associated with particular families. Human experts on
kamon need, of course, to be familiar with the various motifs, and understand
the modifiers that are possible in the system. A typical
manual, such as \cite{Takasawa:21}, will list the motifs with a few dozen
illustrations of each, and will give a few examples of the various modifiers
that are allowed. Humans can fairly easily learn the latter in
most cases with just a few examples, or even one: see
Section~\ref{subsec:human_vs_llm}. For a machine learning system, this same
sparsity makes the task a test of whether motifs, containers, and modifiers
are represented as reusable visual factors, rather than as memorized whole
crest-description pairs.

Our benchmark, described in the next section, consists of synthetic kamon data,
generated using a grammar that incorporates a subset of the combination rules of
kamon. Why use synthetic data rather than, for example, data from already widely
used crests? One reason is that if one is
testing LLMs' abilities at the task of interpreting kamon images into KDL, one
would like a dataset that contains examples that the LLM has likely not seen.
Since we do not know what web sites proprietary LLMs have used for training, using images similar to those found on various sites is not
likely to truly test the LLMs' knowledge of the domain, since they could well
have simply memorized the examples. Synthetic data also gives us access to the
underlying generative factors, which makes it possible to ask whether those
factors are easily recovered in model outputs and internal representations.

\section{The dataset}
\label{sec:dataset}

\begin{figure}[t]
\centering
\begin{tabular}{cccccc}
a) \includegraphics[width=1.5cm]{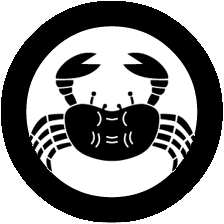} &
b) \includegraphics[width=1.5cm]{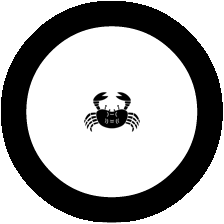} &
c) \includegraphics[width=1.5cm]{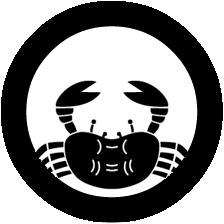} &
d) \includegraphics[width=1.5cm]{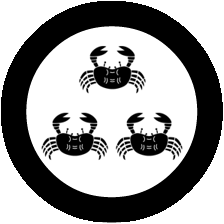} &
e) \includegraphics[width=1.5cm]{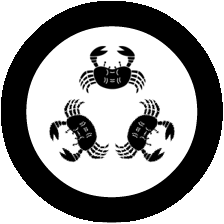} &
f) \includegraphics[width=1.5cm]{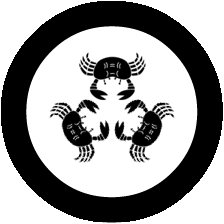} \\
\end{tabular}
\caption{Synthetic examples of crests with various modifiers: a) crab in a
  circle (丸に蟹); b) \emph{bean} crab in a circle (丸に\underline{豆}蟹); c)
  \emph{peeking} crab in a circle (丸に\underline{覗き}蟹); d) \emph{three
  stacked} crabs in a circle (丸に\underline{三つ盛り}蟹); e) \emph{three
  bottoms-together} crabs in a circle (丸に\underline{尻合せ三つ}蟹); f)
  \emph{three heads-together} crabs in a circle (丸に\underline{頭合せ三つ}蟹).}
\label{fig:synth_examples}
\end{figure}

Kamon motifs are divided into two types, \emph{containers} and other
\emph{motifs}. We analyze each composite crest using three factors: an optional
container $C$, a modifier $R$, and a base motif $M$. Using simple image
manipulation, motifs may receive one of three \emph{spatial arrangements}: 三つ盛り
`three-stacked'; 尻合せ三つ `three~bottoms-together'; or 頭合せ三つ
`three~heads-together'. Simple or arranged motifs may also be placed within
containers. In that case, two additional \emph{modifications} are relevant: 豆
`bean' (reduced in size); and 覗き `peeking', i.e., out of the bottom of the
container.
Figure~\ref{fig:synth_examples} shows some examples of these modifiers using
the container 丸 `circle' and the motif 蟹 `crab'. Supplementary Materials~\ref{subapp:bnf}
shows a BNF grammar for the generation process. Note that while the generator
supports multiple levels of containment, the released benchmark does not use
them: a composite example has either one container or no container.

KamonBench contains 20,000 composite examples. Each composite example consists
of a rendered crest, a KDL description, a segmented Japanese analysis of that
description, an English translation, and a non-linguistic program label. The
Japanese analysis tokenizes the KDL description into Japanese-language parts;
the program label records the corresponding generator factors used for
factor-aware evaluation. Across the dataset there are 3,513 possible base
motifs, 36 possible containers, and six
program-label modifier values: null/unmodified, bean, peeking, three-stacked,
bottoms-together, and heads-together.  In the released composite data,
containerless examples use one of the three spatial arrangements; the
null/unmodified value occurs for motifs placed directly inside a container.

The dataset also includes auxiliary component examples. A component example is
a standalone rendered image of a factor used to generate a composite example:
one isolated base motif for every composite example, and one isolated container
for every composite example that uses a container. Thus the dataset has 20,000
composite examples, 20,000 base-motif component examples, and 14,116 container
component examples, for 54,116 examples in total. The 20,000 composite examples
are divided into a 0.8/0.1/0.1 train/development/test split; the corresponding
component examples are included in the same split and used for training. They
are also available for component-level checks. We release three recombination
splits over the composite examples: $(C,M)$, $(R,M)$, and $(C,R,M)$. The
controlled container-motif split uses the same 16,000/2,000/2,000 composite
train/development/test sizes as the main split, with 12,918/1,636/1,656
distinct $(C,M)$ groups. Section~\ref{subsec:cm_controlled} describes these
splits and reports results on all three holdouts.

The component images originate from the \emph{Rebolforces kamondataset},
collected in 2021 from a 2014 archived copy of a former website called Kamon Market. 
The component images originate from the Rebolforces kamondataset, collected in 2021 from a 2014 archived copy of a former website called Kamon Market. We were unable to trace ownership of the IP and thus we could not establish permission to redistribute the renderings. Our release therefore contains annotations, split definitions, per-record provenance metadata and code, excluding the upstream images and dependent composites. Users obtain the upstream archive separately and reconstruct the benchmark locally; image-pixel hashes verify the reconstruction. The release is available at \url{https://github.com/anon-submittere/KamonBench}. Code is licensed under MIT; CC-BY-NC 4.0 applies only to author-created annotations and metadata.

\section{Evaluation enabled by the dataset}
\label{sec:dataset_support}

\paragraph{Factors and grammar mapping.}
KamonBench is designed for factor-aware evaluation of image-to-structure
prediction.  Because each synthetic image is generated from known symbolic
choices, each composite example can be represented as
\[
  Y = (C, R, M),
\]
where $C$ is either a single container or null, $R$ records the modifier
applied to the motif, and $M$ is the base motif.  For contained composites,
$R$ may be null/unmodified, a containment modification, or a spatial
arrangement.  For containerless composites in the released benchmark, $R$ is
one of the three spatial arrangements. The released dataset contains no
recursive containment, so a single triple $(C,R,M)$ represents every composite
example. Supplementary Materials~\ref{subapp:bnf} gives the full BNF grammar. The factor
metrics below are computed on composite examples; component examples are
included in the same splits for training.

\paragraph{Target label spaces.}
The experiments use three target label spaces over the same images and splits.
The Japanese target is a segmented KDL analysis: a Japanese-language sequence
that separates the lexical parts of the KDL description and serves as the
Japanese sequence target. The English target is a translation sequence. The
annotation pipelines for these two target spaces are not released. The
\emph{program} target is derived deterministically from the generator
components, using non-linguistic codes for the container, modifier, and motif.
This makes the $(C,R,M)$ factor metrics direct.
For the ViT and VGG architectures, we train separate baselines for
the Japanese, English, and program targets. This design lets us compare
label-space effects while keeping the image distribution fixed.

\paragraph{Aggregate accuracy and edit distance.}
Aggregate label-space metrics are reported for all three targets on held-out
composite examples. Before computing string-level metrics, whitespace is removed
and katakana is converted to hiragana. Acc is the fraction of examples whose
normalized prediction matches the normalized target. Acc$_{\rm NIT}$ is the
same accuracy restricted to composite test examples for which neither the
target nor the prediction appears in the training set under the corresponding
label-space normalization. CER and TER are both total Levenshtein edit distance
divided by total target length, using characters for Japanese and English and
program tokens for program-code targets. For composite program codes, contained
examples are emitted as three factor tokens, \texttt{C:\ldots},
\texttt{X:\ldots}, and \texttt{M:\ldots}; containerless examples omit the
\texttt{C:\ldots} token.

\paragraph{Controlled recombination.}
Recombination splits hold out combinations of generator factors rather than
individual factors. In an $(R,M)$ split, selected modifier-motif pairs are
assigned to the test examples. The motif in such a pair still
appears in training with other modifiers, and the modifier still appears in
training with other motifs. In a $(C,M)$ split, the same construction is applied
to container-motif pairs: every test composite contains a container-motif
pairing absent from the training composites, while the corresponding container
and motif each occur in other training composites. In a $(C,R,M)$ split, the
full container-modifier-motif combination is held out. The held-out unit is
therefore a combination of familiar factors. Evaluation on these composites asks
whether a model can bind observed visual primitives under new combinations,
rather than relying on frequent complete descriptions.

\paragraph{Counterfactual motif sensitivity.}
The decoder is not constrained by the grammar, so a generated program-code
sequence could omit the modifier or motif code, or emit more than one modifier
or motif code. We call a prediction parseable when it contains optional
container codes, exactly one modifier code, and exactly one motif code. For
a fixed container-modifier context $o=(C,R)$ among contained composites, let
\[
  G_o=\{i: C_i=C,\ R_i=R\}
\]
be the examples with that same container and modifier.  Let $P$ be the set of
eligible pairs $(i,k)$ drawn from the same $G_o$ with different target motifs,
$M_i\neq M_k$, and let $p_i$ indicate that prediction $i$ is parseable.
Consider the following three metrics. Motif separation measures sensitivity to
motif changes:
\[
  \mathrm{Motif\/Separation}=
  \frac{1}{|P|}\sum_{(i,k)\in P}
  \mathbf{1}\{p_i \land p_k \land \hat{M}_i\neq\hat{M}_k\},
\]
which asks whether two examples that differ in target motif also receive
different predicted motif codes.  A score of 1 means that every eligible pair
is separated in the predictions; a score of 0 means that no eligible pair is
separated, for example because the model predicts the same motif throughout the
container-modifier context.  This metric does not require the predicted motifs
to be correct.

Pair motif accuracy adds a correctness requirement:
\[
  \mathrm{PairMotifAcc}=
  \frac{1}{|P|}\sum_{(i,k)\in P}
  \mathbf{1}\{p_i \land p_k \land \hat{M}_i=M_i
  \land \hat{M}_k=M_k\},
\]
which requires both predictions in the pair to contain the correct target
motif.  A score of 1 means that every eligible pair has both motifs correct; a
score of 0 means that no eligible pair has both motifs correct.  When
motif separation is high but pair motif accuracy is low, the model reacts to
motif changes but maps them to the wrong motif identities.

Finally, with $\hat{\mathcal{M}}_o=\{\hat{M}_i : i\in G_o,\ p_i\}$ and
$\mathcal{O}$ the set of evaluated container-modifier contexts,
\[
  \mathrm{CollapsedMotifGroups}=
  \frac{1}{|\mathcal{O}|}\sum_{o\in\mathcal{O}}
  \mathbf{1}\{|\hat{\mathcal{M}}_o|\leq 1\}.
\]
This group-level score detects whether a fixed container-modifier context is
collapsed to a single predicted motif. An individual group contributes 1 when
its parseable predictions contain zero or one distinct motif code, and
contributes 0 when they contain two or more distinct motif codes. Thus an
aggregate score of 1 means that every evaluated container-modifier context is
collapsed, while a score of 0 means that none is collapsed.

\paragraph{Linear representation probes.}
Given a frozen representation $z_i=f_\theta(x_i)$, we train a separate linear
probe for each factor $j\in\{C,R,M\}$,
\[
  q_{\phi,j}(y\mid z)=\mathrm{softmax}(W_j z+b_j),
\]
and evaluate whether the corresponding factor is linearly accessible.  For a
test slice $S$, motif cross-entropy is
\[
  \mathrm{MCE} =
  -\frac{1}{|S|}\sum_{i\in S}\log q_{\phi,M}(M_i\mid z_i),
\]
computed over examples whose motif label is present in the probe training
vocabulary.

KamonBench follows the diagnostic tradition of compositional-generalization
benchmarks such as SCAN and CLEVR
\citep{Lake:Baroni:18,Johnson:EtAl:17}, but casts the problem as visual
recognition in a culturally grounded formal language. The benchmark does not
claim unsupervised disentanglement or identifiable separated latent dimensions.
Instead, it provides supervised factor-recovery and linear-accessibility
diagnostics for known generator factors \citep{Locatello:EtAl:19}.
Supplementary Materials~\ref{subapp:related_work} gives further background on compositional
generalization, factor recovery, probing, and related work on disentanglement
and causal representation learning.

\section{Baselines}
\label{sec:baselines}

\subsection{Baseline models}
\label{subsec:base_models}

We first evaluate three baseline families. The first uses a Vision
Transformer \citep{Dosovitskiy:21} with an autoregressive Transformer decoder.
The second and third use an ImageNet-initialized VGG feature extractor
\citep{Simonyan:Zisserman:15,Mishra:EtAl:21} with an n-gram decoder, either
with learned position-dependent masks or without masks. The masked variant uses
learned image masks so that each output position receives a position-specific
view of the crest from outside to inside. Architecture, training, parameter
counts, and learned-mask examples are given in the
Supplementary Materials~\ref{subapp:baseline_details} and \ref{subapp:masks}.
ViT provides a standard high-capacity image-to-sequence baseline, while the two
VGG variants test whether a simpler convolutional
encoder can exploit, or do without, an explicit positional bias for reading
crests from outer container to inner motif.

\subsection{Baseline results}
\label{sec:base_results}

\begin{table}[t]
\centering
\caption{\label{tab:results}Baseline performance on the 2,000 composite test
  examples. Gray brackets give 95\% bootstrap intervals from full test-set
  resampling with replacement.}
\newcommand{\mci}[2]{\begin{tabular}[t]{@{}c@{}}#1\\[-0.2ex]{\scriptsize\textcolor{gray}{#2}}\end{tabular}}
\begin{tabular}{@{}llccc@{}}
\toprule
Model & Label & CER/TER & Acc & Acc$_{\rm NIT}$\\
\midrule
ViT & Japanese & \mci{0.035}{[0.030, 0.040]} & \mci{0.895}{[0.882, 0.908]} & \mci{0.893}{[0.878, 0.908]}\\
    & English & \mci{0.117}{[0.109, 0.126]} & \mci{0.571}{[0.548, 0.592]} & \mci{0.516}{[0.492, 0.540]}\\
    & Program & \mci{0.022}{[0.018, 0.026]} & \mci{0.941}{[0.931, 0.951]} & \mci{0.945}{[0.934, 0.955]}\\
\midrule
VGG masks & Japanese & \mci{0.161}{[0.152, 0.169]} & \mci{0.498}{[0.476, 0.521]} & \mci{0.446}{[0.421, 0.470]}\\
    & English & \mci{0.208}{[0.198, 0.218]} & \mci{0.398}{[0.377, 0.419]} & \mci{0.312}{[0.289, 0.335]}\\
    & Program & \mci{0.068}{[0.062, 0.075]} & \mci{0.821}{[0.805, 0.836]} & \mci{0.841}{[0.823, 0.858]}\\
\midrule
VGG no masks & Japanese & \mci{0.153}{[0.144, 0.162]} & \mci{0.560}{[0.536, 0.581]} & \mci{0.529}{[0.501, 0.554]}\\
    & English & \mci{0.208}{[0.199, 0.218]} & \mci{0.398}{[0.376, 0.419]} & \mci{0.333}{[0.310, 0.355]}\\
    & Program & \mci{0.101}{[0.094, 0.109]} & \mci{0.732}{[0.711, 0.751]} & \mci{0.783}{[0.761, 0.802]}\\
\bottomrule
\end{tabular}
\end{table}

The ViT and VGG baselines are trained on the 16,000-composite training split
and evaluated on the 2,000-composite test split. They use three target
representations: segmented Japanese analysis labels, English translations, and
non-linguistic program codes. Table~\ref{tab:results} shows that ViT is strongest in all
three label spaces, especially on program codes. The two VGG controls are
close on English, while no-mask VGG is stronger on Japanese and masked VGG is
stronger on program codes.

For program-code outputs, we also map predictions to $(C,R,M)$ factors. We do
not apply this decomposition to Japanese or English outputs, where mapping
natural-language strings back to generator factors would introduce ambiguity.

\begin{table}[t]
\centering
\caption{\label{tab:factor_results}Program-label metrics on the test examples.
  `C Acc' is evaluated only when a container is present; `R' includes
  spatial arrangements and modifications.}
\begin{tabular}{@{}llrrrrr@{}}
\toprule
Model & Slice & N & Acc & C Acc & R Acc & M Acc\\
\midrule
ViT & All & 2000 & 0.941 & 1.000 & 0.994 & 0.946\\
    & Contained & 1431 & 0.932 & 1.000 & 0.992 & 0.938\\
    & Containerless & 569 & 0.963 & \textsc{n/a} & 0.998 & 0.965\\
\midrule
VGG masks & All & 2000 & 0.821 & 0.999 & 0.989 & 0.826\\
    & Contained & 1431 & 0.837 & 0.999 & 0.984 & 0.844\\
    & Containerless & 569 & 0.780 & \textsc{n/a} & 1.000 & 0.780\\
\midrule
VGG no masks & All & 2000 & 0.732 & 0.999 & 0.990 & 0.735\\
    & Contained & 1431 & 0.765 & 0.999 & 0.988 & 0.769\\
    & Containerless & 569 & 0.647 & \textsc{n/a} & 0.996 & 0.649\\
\bottomrule
\end{tabular}
\end{table}

Table~\ref{tab:factor_results} shows that aggregate program accuracy is mainly
a test of motif binding rather than output syntax: all models recover
containers and modifiers nearly perfectly, while contained-motif accuracy
separates ViT, masked VGG, and no-mask VGG.

\begin{table}[t]
\centering
\caption{\label{tab:counterfactual_motif_sensitivity}Counterfactual
motif-sensitivity metrics on the test examples. Each pair shares $(C,R)$ and
differs in motif.}
\begin{tabular}{@{}lrrr@{}}
\toprule
Model & Motif sep. & Pair motif acc. & Collapsed motif groups\\
\midrule
ViT & 1.000 & 0.929 & 0.000\\
VGG masks & 1.000 & 0.870 & 0.000\\
VGG no masks & 1.000 & 0.754 & 0.000\\
\bottomrule
\end{tabular}
\end{table}

Table~\ref{tab:counterfactual_motif_sensitivity} reports motif sensitivity and
pairwise motif accuracy: all models react to motif changes, but pairwise motif
accuracy preserves the same ViT, masked VGG, no-mask VGG ranking.

\subsection{Qwen3-VL baselines}
\label{subsec:qwen}

We also evaluate Qwen3-VL-4B and Qwen3-VL-8B, fine-tuned on the same training
images and program-code targets as the ViT and VGG baselines, including the
auxiliary component examples. Fine-tuning uses low-rank adaptation (LoRA) in
the attention layers; selected vision blocks, the multimodal merger, and the
output head are also trained. Both models are trained for a maximum of 3 or
12 epochs. For the 4B model, we also compare the original tokenizer with an
extended vocabulary containing one token per factor code. Checkpoints are
selected by TER on development composites.
Training settings and details of the output vocabulary are given in the
Supplementary Materials~\ref{subapp:qwen_training}.

\begin{table}[t]
\centering
\caption{\label{tab:qwen}Program-label performance on the 2,000 standard-test
  composites, using seed 20260428. The maximum number of training epochs is
  shown; checkpoints are selected on development data. Across three seeds,
  Qwen3-VL-4B with native tokens and a maximum of 3 epochs achieves TER
  $0.172 \pm 0.002$ and Acc $0.555 \pm 0.007$ (mean $\pm$ SD). Other Qwen
  configurations use one seed. Confidence intervals are reported in
  Supplementary Materials~\ref{subapp:uncertainty}.}
\begin{tabular}{@{}lrlrrr@{}}
\toprule
Model & Max. epochs & Output vocabulary & TER & Acc & M Acc\\
\midrule
Qwen3-VL-4B & 3 & Native & 0.174 & 0.549 & 0.558\\
Qwen3-VL-4B & 12 & Native & 0.079 & 0.796 & 0.804\\
Qwen3-VL-4B & 3 & Factor & 0.071 & 0.814 & 0.826\\
Qwen3-VL-4B & 12 & Factor & 0.031 & 0.918 & 0.924\\
Qwen3-VL-8B & 3 & Native & 0.153 & 0.606 & 0.621\\
Qwen3-VL-8B & 12 & Native & 0.069 & 0.817 & 0.830\\
\midrule
ViT (retrained) & --- & Factor & 0.023 & 0.941 & 0.947\\
\bottomrule
\end{tabular}
\end{table}

Table~\ref{tab:qwen} shows that longer training and factor tokens improve
program accuracy more than increasing model size with native tokens. The best
Qwen configuration achieves an accuracy of 0.918, compared with 0.941 for a
retrained ViT baseline. As with the ViT and VGG models, container and modifier
accuracy remain high, while most errors concern motif identity.

\subsection{Controlled recombination}
\label{subsec:cm_controlled}

We evaluate the $(C,M)$, $(R,M)$, and $(C,R,M)$ splits described in
Section~\ref{sec:dataset_support}. Each split holds out combinations of factors
while retaining the individual factors in training.
Table~\ref{tab:cm_controlled} reports program-code TER and accuracy for ViT,
both VGG variants, and Qwen3-VL-4B. Factor accuracies for the single-run
$(C,M)$ evaluation are given in Table~\ref{tab:recombination_factors}.

\begin{table}[t]
\centering
\caption{\label{tab:cm_controlled}Program-label TER and accuracy on the three
  recombination splits. Each entry gives TER / Acc. ViT results are means over
  three seeds; VGG and Qwen results use one seed. Qwen3-VL-4B uses native tokens
  and a maximum of 3 training epochs. Confidence intervals for all entries
  except VGG on $(C,M)$ are reported in
  Supplementary Materials~\ref{subapp:uncertainty}.}
\begin{tabular}{@{}lccc@{}}
\toprule
Model & $(C,M)$ & $(R,M)$ & $(C,R,M)$\\
\midrule
ViT & 0.030 / 0.920 & 0.038 / 0.900 & 0.026 / 0.931\\
VGG masks & 0.109 / 0.711 & 0.157 / 0.594 & 0.088 / 0.770\\
VGG no masks & 0.160 / 0.572 & 0.257 / 0.330 & 0.112 / 0.709\\
Qwen3-VL-4B & 0.213 / 0.444 & 0.284 / 0.277 & 0.195 / 0.494\\
\bottomrule
\end{tabular}
\end{table}

The $(R,M)$ split is the most difficult for all four models, while accuracy is
highest on $(C,R,M)$. ViT performs best on each split, and masked VGG
outperforms VGG without masks. Container and modifier accuracy remain high
across the three splits, with most errors attributable to motif identity.

In the Supplementary Materials~\ref{subapp:vgg_failure}, we report results on an
initial VGG variant featuring a program-label failure that motivated these
diagnostics.

\subsection{Representation-level factor accessibility}
\label{subsec:representation_accessibility}

The output diagnostics above ask whether generated descriptions bind the
correct factors. We also train linear probes on frozen representations from the
program-code baselines to test whether $C$, $R$, and $M$ are separately
accessible before decoding. For VGG, probes use the concatenation of the
per-position VGG features passed to the n-gram decoder. For ViT, probes use the
mean-pooled encoder features passed from the ViT image encoder to the Transformer
decoder.
Probe accuracy is a limited representation test: it shows what a linear readout
can recover from the frozen feature, not whether the decoder actually uses that
information or whether the information is available nonlinearly.

\begin{table}[t]
\centering
\caption{\label{tab:representation_probes}Linear probes on frozen program-model
  representations. Acc columns report factor probe accuracies; motif accuracy is
  reported overall, on contained examples, and on containerless examples. M CE is
  motif cross-entropy.}
\begin{tabular}{@{}lrrrrrr@{}}
\toprule
Model & C Acc & R Acc & M Acc & M Acc$_{\mathrm{cont}}$ &
M Acc$_{\mathrm{noC}}$ & M CE\\
\midrule
ViT & 1.000 & 0.988 & 0.776 & 0.730 & 0.893 & 1.401\\
VGG masks & 0.999 & 0.987 & 0.552 & 0.599 & 0.434 & 4.905\\
VGG no masks & 0.999 & 0.988 & 0.412 & 0.430 & 0.367 & 6.520\\
\bottomrule
\end{tabular}
\end{table}

Table~\ref{tab:representation_probes} shows that container and modifier labels
are almost linearly accessible in all three representations, while motif
identity is less accessible in the VGG representations.
Table~\ref{tab:probe_support} further shows that motif
accessibility drops on factor combinations absent from training, with ViT
remaining strongest.

\begin{table}[t]
\centering
\caption{\label{tab:probe_support}Motif-probe accuracy on standard-test examples
  whose factor combinations are absent or present in the training split.}
\begin{tabular}{@{}llrrrr@{}}
\toprule
Model & Factor combination & N absent & Acc absent & N present & Acc present\\
\midrule
ViT & $(C,M)$ & 1450 & 0.702 & 550 & 0.971\\
    & $(R,M)$ & 607 & 0.554 & 1393 & 0.873\\
    & $(C,R,M)$ & 1698 & 0.736 & 302 & 1.000\\
\midrule
VGG masks & $(C,M)$ & 1450 & 0.519 & 550 & 0.638\\
    & $(R,M)$ & 607 & 0.091 & 1393 & 0.753\\
    & $(C,R,M)$ & 1698 & 0.488 & 302 & 0.914\\
\midrule
VGG no masks & $(C,M)$ & 1450 & 0.366 & 550 & 0.535\\
    & $(R,M)$ & 607 & 0.099 & 1393 & 0.548\\
    & $(C,R,M)$ & 1698 & 0.355 & 302 & 0.735\\
\bottomrule
\end{tabular}
\end{table}

\subsection{Few-shot multimodal LLM performance}
\label{subsec:llm}

We provided Claude~Opus~4.7~Max and GPT~5.4~xhigh with five labeled kamon
examples and asked them to describe 20 randomly sampled synthetic test images
in Japanese, without fine-tuning on KamonBench. The prompt is shown in the Supplementary
Materials~\ref{subapp:llm_prompt_20ex}, and the resulting outputs in
Table~\ref{tab:llms}. On these examples, the masked VGG baseline makes one error and
ViT makes none. Claude produces no correct transcriptions, and GPT
produces one, although both models sometimes recover individual components
such as a container or motif. This experiment tests a practical confound for
kamon evaluation: large proprietary multimodal models may have seen kamon
images and KDL descriptions on the web. The results in Table~\ref{tab:llms}
indicate that such exposure, if present, is not sufficient for LLMs to succeed
in a few-shot setting on our synthetic benchmark.

\subsection{Few-example human and LLM evaluation}
\label{subsec:human_vs_llm}

We also evaluated whether non-expert humans can learn local aspects of the
kamon construction system from a small amount of instruction. Participants first
received descriptions of several basic motifs, modifications, and spatial
arrangements. They were then presented with ten synthetic examples and asked to
identify, for each example, the basic motif, any modification (e.g. 鬼 `demon',
重ね `overlapping'), and any spatial arrangement (e.g. 三つ盛り `stacked',
尻合わせ `bottoms together'). Participants could select instructions in English
(Supplementary Materials~\ref{subapp:human_instr_en}) or Japanese
(Supplementary Materials~\ref{subapp:human_instr_ja}). A sample Google Form question from the
questionnaire is shown in the Supplementary Materials~\ref{subapp:human_quest}. In addition to
questions about the crests, participants were asked to self-report their level
of knowledge of kamon, on a 1--4 scale, where 1 is `no knowledge' and 4 is
`expert'.  The task took around 10 minutes for the participants who took part.
There were 32 participants.
This task is complementary to the KDL transcription prompts: it isolates
whether names and operations can be learned from a short tutorial, while the
model experiments ask whether those factors can be recombined into structured
outputs.

Participants exhibited a wide range of numbers of errors, ranging from no errors
for 2 participants, to 10 errors for 1. The largest group of errors related to
modification (64\%). Most (22) participants reported no knowledge of kamon (1),
with the remainder reporting some knowledge (2). There was no
significant difference in performance between these two groups.  In fact, the 2
participants who made no errors both self-reported as having no knowledge (1),
whereas the one participant who made 10 errors self-reported as having some
knowledge (2).

\begin{figure}[h]
\centering
\includegraphics[width=0.60\textwidth]{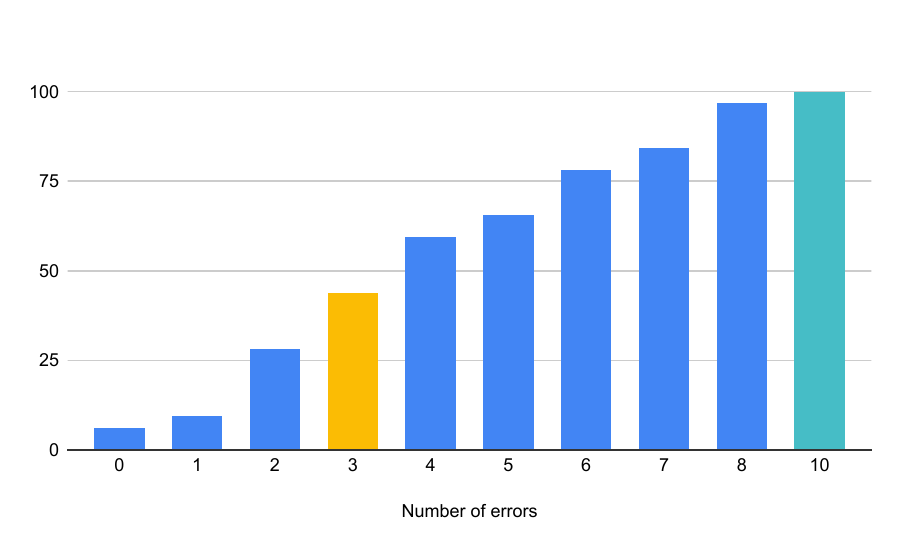}
\caption{Cumulative error counts for the 32 participants, ranked from the
  best (2 with no errors) to the worst (1 with 10 errors). Orange bar:
  GPT~5.4~xhigh; turquoise bar: Claude~Opus~4.7~Max. 100\% of participants
  did at least as well as Claude; 44\% at least as well as GPT.}
\label{fig:cumulative_errors}
\end{figure}

We prompted GPT~5.4~xhigh and Claude~Opus~4.7~Max with the same ten-example
task, using the prompt shown in the Supplementary Materials~\ref{subapp:llm_prompt_10}. Performance
again varied: GPT made 3 errors, whereas Claude made 10 errors. Modification
identification was also the hardest category for the two LLMs, accounting for all of
GPT's errors and 60\% of Claude's errors. All human participants performed at
least as well as Claude. Relative to GPT, 28\% of participants
made 2 or fewer errors and
therefore outperformed GPT, while 44\% made 3 or fewer errors and therefore
matched or outperformed GPT. Figure~\ref{fig:cumulative_errors} plots the
distribution.

Kamon analysis is challenging even for human participants, but the human results
show that some aspects of the construction system can be acquired with minimal
instruction: 44\% of participants matched or exceeded the best LLM performance
observed here. If participants' self-reported prior knowledge is accurate, this
performance reflects learning from the provided examples rather than prior
formal training in kamon. For proprietary models, by contrast, prior exposure to
kamon data cannot be audited.

\section{Conclusion and limitations}
\label{sec:summary}
\label{sec:limitations}

KamonBench uses kamon, a Japanese heraldic tradition with centuries of
historical use and a specialized descriptive vocabulary, as a benchmark for
image-to-structure prediction. Its motivation is twofold: kamon provide a
culturally grounded formal language with meaningful containers, modifiers, and
motifs, while the synthetic generator exposes those units as known factors,
supporting controlled tests of sparse recombination.

The experiments show that this factor-aware view changes what the benchmark
measures. Aggregate string metrics identify ViT as the strongest
baseline, but program labels reveal that all baselines mostly recover
containers and modifiers while motif identity under containment remains the main
bottleneck. Longer training and factor tokens improve Qwen3-VL performance,
but do not close the gap to ViT. Controlled $(C,M)$, $(R,M)$, and $(C,R,M)$
recombination, counterfactual motif-sensitivity tests, and linear probes
separate local factor recognition, compositional
binding in outputs, and factor accessibility in frozen representations. The
few-example human and multimodal LLM studies further motivate the setting:
local aspects of KDL can be learned from limited instruction, while closed
proprietary models do not reliably solve the synthetic task from a small prompt.

The benchmark's control also limits its scope. Generated crests are less
polished than professionally rendered kamon and differ from real crests in
books or on the web. The supplementary evaluation on 114 real images shows
poor transfer from synthetic training
(Supplementary Materials~\ref{subapp:real_transfer}). We evaluate a small set of baselines,
leave Japanese and English outputs as strings rather than mapping them back into
generator factors, and probe only linear accessibility from one feature vector
per image. We stress however that while ML methods show poor transfer to professionally rendered
kamon, due to the rendering quality issues noted above, such differences have significantly less effect
for human raters. A human judge would likely be able to learn the modifications shown in Figure~\ref{fig:synth_examples}, and 
apply them to professionally rendered kamon, without being distracted by the rendering quality differences of the
synthetic examples. This underscores an important point that current vision models are still distant in 
their behavior from human vision, and we offer KamonBench as a well-controlled benchmark for 
exploring closing this gap. Pursuing this research direction is left to future research.



\newpage

\bibliography{refs}
\bibliographystyle{tmlr}

\newpage

\appendix
\raggedbottom
\setlength{\intextsep}{8pt plus 2pt minus 2pt}

\renewcommand{\thesection}{Supplementary materials}
\renewcommand{\thesubsection}{S.\arabic{subsection}}

\newcommand{\documentimage}[3]{%
  \includegraphics[page=#2,trim=#3,clip,scale=0.8]{Figures/#1.pdf}%
}
\newcommand{\appendiximage}[4]{%
  \par\noindent\makebox[\textwidth]{%
    \documentimage{#1}{#2}{60bp #3bp 60.2756bp #4bp}}\par
}

\section{}

\subsection{Background and related work}
\label{subapp:related_work}
\label{sec:factor_recovery}

KamonBench is designed around three labeled factors of variation per crest:
container $C$, modifier $R$, and motif $M$. It provides a suite of
factor-aware diagnostics defined in Section~\ref{sec:dataset_support}. This
section positions those design choices relative to compositional
generalization, factor recovery, linear probing, and related work on
disentanglement and causal representation learning.

\paragraph{Compositional generalization.}
Systematic compositional generalization is usually tested by asking whether a
model can recombine primitives in configurations that were rare or absent during
training.  Text-only benchmarks such as SCAN expose failures to generalize
compositionally even when the primitive vocabulary is known
\citep{Lake:Baroni:18}.  Visual-reasoning benchmarks such as CLEVR use
synthetic scenes and executable annotations to separate reasoning skills that
are confounded in aggregate question-answering accuracy
\citep{Johnson:EtAl:17}.  KamonBench follows this diagnostic tradition and
casts the problem as image-to-structure prediction in a cultural formal
language: the input is a rendered crest, and the target is a structured
description whose factors are known by construction.  In this setting, the
relevant primitives are visually grounded motifs, containers, and modifiers;
composition requires recognizing these elements and binding them into a valid
KDL description.
Hupkes et al.'s taxonomy separates several kinds of compositional
generalization, including systematicity, productivity, localism,
substitutivity, and overgeneralisation \citep{Hupkes:EtAl:20}.  In that
taxonomy, our controlled $(C,M)$ holdout is closest to systematicity: each
container and motif appears in training, while specific pairs of containers and
motifs are withheld, so the test evaluates recombination of known primitives.
Hupkes et al.'s substitutivity test concerns synonym substitution, which is
not the construction we use; instead, our counterfactual motif groups
(Section~\ref{sec:dataset_support},
Table~\ref{tab:counterfactual_motif_sensitivity}) are framed as single-factor
interventions on motif identity under a fixed container-modifier context
$(C,R)$.

\paragraph{Factor recovery and probes.}
In representation learning, disentanglement describes codes in which distinct
explanatory factors of variation are separated
\citep{Higgins:EtAl:17,Higgins:EtAl:18}.  This literature has also shown that
unsupervised disentanglement is not identifiable without appropriate inductive
biases or supervision \citep{Locatello:EtAl:19}. KamonBench does not test
disentanglement in this stronger sense. Instead, because the generator exposes
factor labels by construction, we evaluate supervised factor recovery: whether
outputs and frozen representations recover the labeled container, modifier,
and motif, and whether those factors remain recoverable under recombination.
This framing is consistent with few-label disentanglement work, where limited
factor annotations are used for model selection or semi-supervised training
\citep{Locatello:EtAl:20}.
Our representation-level analyses follow the linear-probing tradition for
studying what information is accessible in learned representations
\citep{Alain:Bengio:17,Hewitt:Manning:19,Belinkov:22}.  The counterfactual
factor tests are also aligned with the causal-representation-learning view that
useful representations should support controlled interventions on high-level
factors \citep{Schoelkopf:EtAl:21}.


\clearpage

\subsection{BNF for synthetic kamon generation}
\label{subapp:bnf}

A BNF grammar that defines the possible factor structures is shown in
Figure~\ref{fig:bnf}.

\begin{figure}[H]
\noindent\rule{\textwidth}{0.4pt}\par\smallskip
\noindent\begin{tabular}{@{}l@{\quad}c@{\quad}l@{}}
  \synt{kamon} & ::= & \synt{contained} \quad\textbar\quad \synt{spatial-arrangement} \synt{MOTIF} \\
  \synt{contained} & ::= & \synt{CONTAINER} \synt{complex-motif} \\
  \synt{complex-motif} & ::= & \synt{contained-modifier} \synt{MOTIF} \quad\textbar\quad \synt{contained} \\
  \synt{contained-modifier} & ::= & \synt{modifier} \quad\textbar\quad \synt{empty} \\
  \synt{modifier} & ::= & \synt{spatial-arrangement} \quad\textbar\quad \synt{modification} \\
  \synt{spatial-arrangement} & ::= & 三つ盛り/three-stacked \\
   & & \textbar\quad 尻合せ三つ/three~bottoms-together \\
   & & \textbar\quad 頭合せ三つ/three~heads-together \\
  \synt{modification} & ::= & 豆/bean \quad\textbar\quad 覗き/peeking
\end{tabular}\par\smallskip
\noindent\rule{\textwidth}{0.4pt}
\caption{BNF for kamon generation. Valid $\langle$\emph{CONTAINER}$\rangle$s and
  $\langle$\emph{MOTIF}$\rangle$s are provided with the released benchmark
  dataset. The $\langle$\emph{modifier}$\rangle$ non-terminal covers both
  spatial arrangements and modifications.  The $\langle$\emph{empty}$\rangle$
  alternative denotes the null/unmodified value for a motif placed directly
  inside a container.  Containerless composite examples use a spatial
  arrangement. Note that the recursion on the
  $\langle$\emph{complex-motif}$\rangle$ node, while supported by the generator,
  is not used in the generation of the current dataset.}
\label{fig:bnf}
\end{figure}

\subsection{Baseline architecture and training details}
\label{subapp:baseline_details}

\begin{figure}[H]
\centering
\includegraphics[width=\textwidth]{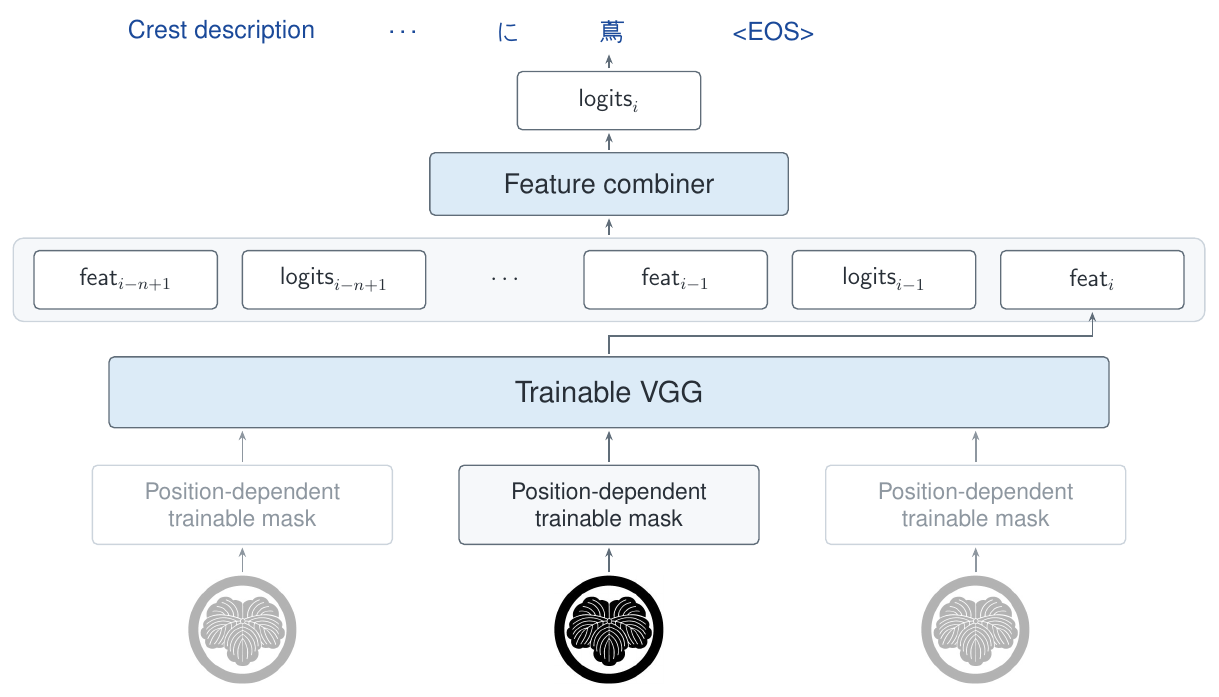}
\caption{Schematic VGG n-gram decoder family. The blue components are shared
  across output positions.}
\label{fig:model}
\end{figure}
\noindent\begin{minipage}{\textwidth}
In the masked VGG n-gram baseline, $I$ is the input image, $P_i$ is the
position-dependent mask, $F_i$ is the feature at position $i$, $L_i$ is the
logit at position $i$, and FC is the MLP feature combiner that maps the
concatenated context $H_i$ to output logits:
\begin{align}
  F_i &= \mathrm{FE}(I \cdot P_i),\nonumber\\
  H_i &= \mathrm{Concat}[
    F_{i-n+1}, L_{i-n+1},
    \ldots,
    F_{i-1}, L_{i-1},
    F_i],\nonumber\\
  L_i &= \mathrm{FC}(H_i).\nonumber
\end{align}
Terms with indices below the first output position are omitted; the n-gram-1
case therefore reduces to $H_i=F_i$.
\end{minipage}\par

Tables~\ref{tab:baseline_architecture}--\ref{tab:baseline_checkpoints} summarize
the architecture, optimization, and selected checkpoints for the baselines used
in Table~\ref{tab:results}.

\begin{table}[H]
\centering
\footnotesize
\caption{\label{tab:baseline_architecture}Architecture hyperparameters for the
  three baseline families. All image backbones are initialized from
  ImageNet-pretrained weights and fine-tuned.}
\setlength{\tabcolsep}{3pt}
\begin{tabular}{@{}p{0.19\textwidth}p{0.23\textwidth}p{0.24\textwidth}p{0.24\textwidth}@{}}
\toprule
Setting & ViT & VGG masks & VGG no masks\\
\midrule
Image encoder & \texttt{vit\_base\_patch16\_224} & VGG16 backbone & VGG16 backbone\\
Image size & 224 & 224 & 224\\
Decoder/interface & 4-layer Transformer decoder & MLP n-gram decoder & MLP n-gram decoder\\
Decoder width & 512 & 512 & 512\\
Attention heads & 8 & \textsc{n/a} & \textsc{n/a}\\
Decoder context & autoregressive tokens & n-gram length 1 & n-gram length 4\\
Position mechanism & ViT patch features & learned image masks & no learned masks\\
Dropout & 0.1 & 0.1 in MLP & 0.1 in MLP\\
Token dropout & 0.0 & \textsc{n/a} & \textsc{n/a}\\
\bottomrule
\end{tabular}
\end{table}

\begin{table}[H]
\centering
\footnotesize
\caption{\label{tab:baseline_training}Training and checkpoint-selection
  settings for the three baseline families.}
\setlength{\tabcolsep}{3pt}
\begin{tabular}{@{}llll@{}}
\toprule
Setting & ViT & VGG masks & VGG no masks\\
\midrule
Optimizer & AdamW & AdamW & AdamW\\
Main/decoder learning rate & $1.5\times10^{-4}$ & $10^{-4}$ & $10^{-4}$\\
Backbone learning rate & $10^{-5}$ & same as main LR & same as main LR\\
Weight decay & $10^{-4}$ & $10^{-4}$ & $10^{-4}$\\
Label smoothing & 0.02 & 0.02 & 0.02\\
Batch size & 64 & 64 & 64\\
Early-stop patience & 10 & 10 & 10\\
Selection metric & validation CER/TER & composite validation error & composite validation error\\
\bottomrule
\end{tabular}
\end{table}

\begin{table}[H]
\centering
\small
\caption{\label{tab:baseline_checkpoints}Selected checkpoints and parameter
  counts for the baselines. Parameter counts are computed from
  checkpoint parameter tensors and exclude non-parameter buffers.}
\begin{tabular}{@{}llrrrr@{}}
\toprule
Family & Label & Selected step & Vocabulary size & Max length & Parameters\\
\midrule
ViT & English & 38,000 & 2,188 & 15 & 105,260,428\\
ViT & Japanese & 78,000 & 1,858 & 12 & 104,920,642\\
ViT & Program & 54,000 & 3,556 & 4 & 106,656,996\\
\midrule
VGG masks & English & 25,000 & 2,188 & 15 & 138,495,948\\
VGG masks & Japanese & 22,500 & 1,858 & 12 & 138,176,130\\
VGG masks & Program & 23,750 & 3,556 & 4 & 138,645,796\\
\midrule
VGG no masks & English & 25,000 & 2,188 & 15 & 147,395,532\\
VGG no masks & Japanese & 28,750 & 1,858 & 12 & 146,719,362\\
VGG no masks & Program & 42,500 & 3,556 & 4 & 150,198,564\\
\bottomrule
\end{tabular}
\end{table}

The reported experiments fit on a single H100 GPU node.

\subsection{Qwen3-VL training details}
\label{subapp:qwen_training}

On the standard split, we fine-tune Qwen3-VL-4B-Instruct and Qwen3-VL-8B-Instruct
on 16,000 composites and 27,280 component records. Each epoch presents the original
image and three augmentations per record. The prompt is ``Return only the
KamonBench program code. Do not explain.'' Cross-entropy loss is applied only
to the target response. Table~\ref{tab:qwen_training} gives the training
settings. Each run starts from pretrained weights, with the cosine schedule
spanning its maximum of 3 or 12 epochs. We evaluate every quarter epoch and
select the checkpoint with the lowest development-composite TER, breaking
ties by higher exact accuracy and then earlier step.

\begin{table}[H]
\centering
\small
\caption{\label{tab:qwen_training}Training and decoding settings for the
  Qwen3-VL baselines.}
\begin{tabular}{@{}p{0.30\textwidth}p{0.64\textwidth}@{}}
\toprule
Setting & Value\\
\midrule
Image size & $224\times224$\\
LoRA & Rank 16, $\alpha=32$, dropout 0.05\\
LoRA projections & Query, key, value, and output in language attention\\
Other trainable modules & Final two vision blocks, multimodal merger, output head\\
Optimizer & AdamW, weight decay 0\\
Learning rates & LoRA: $10^{-4}$; merger/head: $10^{-5}$; vision: $2\times10^{-6}$\\
Global batch size & 64\\
Schedule & Cosine decay, 3\% warmup\\
Gradient clipping & Maximum norm 1\\
Precision & FP32 trainable weights; BF16 frozen weights and autocast\\
Decoding & Greedy, at most 24 new tokens, no grammar constraint\\
\bottomrule
\end{tabular}
\end{table}

For factor-token outputs, we add 3,555 vocabulary entries, one per complete
factor code. New entries are initialized from the mean input embedding of
their native-token sequences; input embeddings remain frozen, while the
expanded output head is trained. Spaces retain their native tokens. This
adds 8,417,280 trainable output-head parameters to the 4B model.

\paragraph{Isolated and composite motifs.}
Qwen3-VL-4B with native tokens and three training epochs achieves motif
accuracy of 0.932 on isolated motifs and 0.558 on the corresponding
composites. The retrained ViT achieves 0.987 and 0.947, respectively.

\subsection{Uncertainty estimates}
\label{subapp:uncertainty}

Tables~\ref{tab:qwen_intervals} and \ref{tab:recombination_intervals} report
95\% percentile bootstrap intervals from 10,000 resamples of test images,
holding the trained checkpoints fixed. For ViT's three-seed means, the same
images are resampled across runs. Standard deviations across training seeds
are reported separately.

\begin{table}[H]
\centering
\small
\caption{\label{tab:qwen_intervals}95\% confidence intervals for the
  standard-split results in Table~\ref{tab:qwen}.}
\begin{tabular}{@{}lrlccc@{}}
\toprule
Model & Max. epochs & Vocabulary & TER & Acc & M Acc\\
\midrule
Qwen3-VL-4B & 3 & Native & [0.165, 0.183] & [0.528, 0.571] & [0.536, 0.580]\\
Qwen3-VL-4B & 12 & Native & [0.072, 0.086] & [0.778, 0.814] & [0.787, 0.822]\\
Qwen3-VL-4B & 3 & Factor & [0.065, 0.078] & [0.797, 0.831] & [0.809, 0.842]\\
Qwen3-VL-4B & 12 & Factor & [0.027, 0.036] & [0.906, 0.929] & [0.912, 0.936]\\
Qwen3-VL-8B & 3 & Native & [0.144, 0.162] & [0.585, 0.628] & [0.600, 0.642]\\
Qwen3-VL-8B & 12 & Native & [0.063, 0.076] & [0.800, 0.834] & [0.813, 0.846]\\
\midrule
ViT (retrained) & --- & Factor & [0.019, 0.027] & [0.930, 0.951] & [0.937, 0.957]\\
\bottomrule
\end{tabular}
\end{table}

\begin{table}[H]
\centering
\small
\caption{\label{tab:recombination_intervals}95\% confidence intervals for
  recombination results. The last column gives ViT's standard deviation across
  three training seeds.}
\begin{tabular}{@{}llccc@{}}
\toprule
Model & Split & TER & Acc & Seed SD (TER / Acc)\\
\midrule
ViT & $(C,M)$ & [0.026, 0.033] & [0.910, 0.930] & 0.001 / 0.004\\
    & $(R,M)$ & [0.034, 0.043] & [0.888, 0.911] & 0.002 / 0.005\\
    & $(C,R,M)$ & [0.022, 0.030] & [0.922, 0.940] & 0.003 / 0.007\\
\midrule
VGG masks & $(R,M)$ & [0.148, 0.166] & [0.573, 0.616] & ---\\
          & $(C,R,M)$ & [0.081, 0.095] & [0.751, 0.788] & ---\\
VGG no masks & $(R,M)$ & [0.249, 0.266] & [0.309, 0.350] & ---\\
             & $(C,R,M)$ & [0.104, 0.120] & [0.688, 0.728] & ---\\
\midrule
Qwen3-VL-4B & $(C,M)$ & [0.204, 0.222] & [0.422, 0.466] & ---\\
            & $(R,M)$ & [0.276, 0.292] & [0.258, 0.296] & ---\\
            & $(C,R,M)$ & [0.186, 0.204] & [0.472, 0.516] & ---\\
\bottomrule
\end{tabular}
\end{table}

\subsection{Recombination factor accuracies}
\label{subapp:recombination_factors}

\begin{table}[H]
\centering
\small
\caption{\label{tab:recombination_factors}Single-run factor accuracies on the
  $(C,M)$ recombination split.}
\begin{tabular}{@{}lrrr@{}}
\toprule
Model & C Acc & R Acc & M Acc\\
\midrule
ViT & 0.999 & 0.994 & 0.912\\
VGG masks & 0.997 & 0.987 & 0.716\\
VGG no masks & 0.997 & 0.992 & 0.572\\
\bottomrule
\end{tabular}
\end{table}

\subsection{Synthetic-to-real transfer}
\label{subapp:real_transfer}

This supplementary evaluation examines generalization beyond the controlled
synthetic setting considered in the main experiments.

\paragraph{Images and targets.}
We screened 1,002 description-linked historical and Wikimedia images and
retained 115 candidates through exact description matching to the benchmark
grammar. A kamon expert reviewed the proposed targets without access to model
predictions, correcting two motif and two container labels and excluding one
image whose motif was unavailable in the target vocabulary. The resulting set
contains 114 images spanning 49 programs: 86 crops from digitized
\emph{Ansei Bukan}, Edo-period printed samurai directories provided by the
Center for Open Data in the Humanities, and 28 Wikimedia renderings.
The evaluation is therefore restricted to images with descriptions expressible
by the benchmark grammar.

\paragraph{Evaluation.}
We evaluate the standard-split checkpoints without further training, using
their original preprocessing. Qwen3-VL-4B uses native tokens and a maximum of
3 training epochs. Each program is paired with one synthetic rendering; the
49 controls are weighted to match the program frequencies among the 114 real
images. Confidence intervals for real-minus-control differences use 10,000
bootstrap samples clustered by target program: each program's real images and
matched control are resampled together, retaining image weighting.

\begin{table}[H]
\centering
\small
\caption{\label{tab:real_transfer}Program-label performance on 114 real kamon
  images and matched synthetic controls with the original preprocessing.
  Differences are real minus control; gray brackets give 95\% program-cluster
  bootstrap intervals for those differences.}
\newcommand{\rtci}[2]{\begin{tabular}[t]{@{}c@{}}#1\\[-0.2ex]{\scriptsize\textcolor{gray}{#2}}\end{tabular}}
\setlength{\tabcolsep}{4pt}
\begin{tabular}{@{}lcccc@{}}
\toprule
Model & Real TER / Acc & Control TER / Acc & $\Delta$TER & $\Delta$Acc\\
\midrule
ViT & 0.792 / 0.000 & 0.021 / 0.939 & \rtci{0.770}{[0.711, 0.826]} & \rtci{$-0.939$}{$[-0.992,-0.856]$}\\
VGG masks & 0.849 / 0.000 & 0.018 / 0.947 & \rtci{0.831}{[0.785, 0.866]} & \rtci{$-0.947$}{$[-1.000,-0.870]$}\\
VGG no masks & 0.831 / 0.009 & 0.045 / 0.868 & \rtci{0.785}{[0.714, 0.847]} & \rtci{$-0.860$}{$[-0.973,-0.701]$}\\
Qwen3-VL-4B & 0.852 / 0.000 & 0.184 / 0.465 & \rtci{0.668}{[0.605, 0.746]} & \rtci{$-0.465$}{$[-0.684,-0.299]$}\\
\bottomrule
\end{tabular}
\end{table}

Table~\ref{tab:real_transfer} shows poor transfer for all four models despite
substantially stronger performance on matched controls. No model recovers an
exact program from the historical crops; VGG without masks correctly predicts
one Wikimedia image. All target factors occur in synthetic training for 111 of
the 114 images. Restricting evaluation to these images leaves TER between
0.789 and 0.848 and accuracy at most 0.009, with the same overall conclusion.

\paragraph{Preprocessing.}
We additionally examine grayscale conversion, foreground binarization, and
binarization with alignment and denoising. Grayscale conversion preserves
polarity. Foreground binarization normalizes polarity using the border median
and applies a global Otsu threshold to obtain black foreground on white.
The aligned variant preserves aspect ratio and additionally applies median
denoising, morphological closing, removal of small connected components, and
centering and scaling on a $224\times224$ canvas. These transformations operate
without target labels. Only real images are transformed; synthetic controls
remain unchanged.

\begin{table}[H]
\centering
\small
\caption{\label{tab:real_preprocessing}Real-image preprocessing sensitivity.
  Each entry gives TER / Acc on the same 114 images. Models are unchanged
  across conditions.}
\begin{tabular}{@{}lcccc@{}}
\toprule
Model & Original & Grayscale & Binary & Aligned binary\\
\midrule
ViT & 0.792 / 0.000 & 0.792 / 0.000 & 0.755 / 0.018 & 0.801 / 0.009\\
VGG masks & 0.849 / 0.000 & 0.843 / 0.000 & 0.858 / 0.000 & 0.888 / 0.000\\
VGG no masks & 0.831 / 0.009 & 0.834 / 0.009 & 0.849 / 0.009 & 0.873 / 0.018\\
Qwen3-VL-4B & 0.852 / 0.000 & 0.876 / 0.000 & 0.704 / 0.000 & 0.792 / 0.000\\
\bottomrule
\end{tabular}
\end{table}

Foreground binarization reduces ViT TER by 0.036 (95\% interval
$[0.003,0.079]$) and Qwen TER by 0.148 ($[0.094,0.201]$), using paired
program-cluster bootstrap intervals. Their accuracies remain 0.018 (two images)
and 0.000, respectively. Further alignment and denoising do not improve either
model over foreground binarization. Simple preprocessing therefore reduces
some errors but does not resolve the observed transfer~gap.

\subsection{Reduced-training-data baseline sweep}
\label{subapp:train_size_sweep}

To measure data scaling under the same baselines, we retrained ViT, masked VGG,
and no-mask VGG on deterministic training subsets containing 2,500, 5,000, or
10,000 composite examples. We evaluate on the standard development and test
splits.

Table~\ref{tab:train_size_results} reports the same aggregate composite-test
metrics as Table~\ref{tab:results}. Table~\ref{tab:train_size_factors} reports
the same program-code factor metrics as
Table~\ref{tab:factor_results}.

\begingroup
\small
\newcommand{\tsci}[2]{\begin{tabular}[t]{@{}c@{}}#1\\[-0.2ex]{\scriptsize\textcolor{gray}{#2}}\end{tabular}}
\begin{longtable}{@{}rllccc@{}}
\caption{\label{tab:train_size_results}Reduced-training-data performance on the standard test data. Train size counts selected training examples. Gray brackets give 95\% bootstrap intervals.}\\
\toprule
Train & Model & Label & CER/TER & Acc & Acc$_{\rm NIT}$\\
\midrule
\endfirsthead
\caption[]{Reduced-training-data performance (continued).}\\
\toprule
Train & Model & Label & CER/TER & Acc & Acc$_{\rm NIT}$\\
\midrule
\endhead
\midrule
\multicolumn{6}{r}{\small Continued on the next page.}\\
\endfoot
\bottomrule
\endlastfoot
2,500 & ViT & Japanese & \tsci{0.273}{[0.262, 0.285]} & \tsci{0.368}{[0.347, 0.390]} & \tsci{0.387}{[0.364, 0.409]}\\*
 &  & English & \tsci{0.303}{[0.293, 0.315]} & \tsci{0.234}{[0.216, 0.249]} & \tsci{0.235}{[0.216, 0.252]}\\*
 &  & Program & \tsci{0.219}{[0.211, 0.227]} & \tsci{0.424}{[0.402, 0.445]} & \tsci{0.477}{[0.453, 0.501]}\\
\addlinespace[0.25ex]
2,500 & VGG masks & Japanese & \tsci{0.387}{[0.378, 0.396]} & \tsci{0.099}{[0.086, 0.112]} & \tsci{0.082}{[0.069, 0.096]}\\*
 &  & English & \tsci{0.413}{[0.403, 0.422]} & \tsci{0.085}{[0.072, 0.097]} & \tsci{0.067}{[0.055, 0.078]}\\*
 &  & Program & \tsci{0.284}{[0.275, 0.293]} & \tsci{0.276}{[0.256, 0.295]} & \tsci{0.370}{[0.343, 0.396]}\\
\addlinespace[0.25ex]
2,500 & VGG no masks & Japanese & \tsci{0.411}{[0.401, 0.420]} & \tsci{0.103}{[0.089, 0.117]} & \tsci{0.097}{[0.082, 0.110]}\\*
 &  & English & \tsci{0.408}{[0.399, 0.419]} & \tsci{0.102}{[0.089, 0.116]} & \tsci{0.093}{[0.079, 0.107]}\\*
 &  & Program & \tsci{0.337}{[0.329, 0.344]} & \tsci{0.146}{[0.132, 0.162]} & \tsci{0.320}{[0.289, 0.352]}\\
\midrule
5,000 & ViT & Japanese & \tsci{0.157}{[0.147, 0.167]} & \tsci{0.600}{[0.580, 0.621]} & \tsci{0.614}{[0.592, 0.635]}\\*
 &  & English & \tsci{0.221}{[0.211, 0.231]} & \tsci{0.359}{[0.339, 0.380]} & \tsci{0.351}{[0.330, 0.373]}\\*
 &  & Program & \tsci{0.117}{[0.109, 0.124]} & \tsci{0.691}{[0.672, 0.710]} & \tsci{0.720}{[0.700, 0.740]}\\
\addlinespace[0.25ex]
5,000 & VGG masks & Japanese & \tsci{0.319}{[0.310, 0.329]} & \tsci{0.203}{[0.185, 0.222]} & \tsci{0.176}{[0.158, 0.194]}\\*
 &  & English & \tsci{0.334}{[0.325, 0.344]} & \tsci{0.170}{[0.153, 0.186]} & \tsci{0.136}{[0.120, 0.151]}\\*
 &  & Program & \tsci{0.199}{[0.190, 0.209]} & \tsci{0.479}{[0.456, 0.500]} & \tsci{0.574}{[0.547, 0.597]}\\
\addlinespace[0.25ex]
5,000 & VGG no masks & Japanese & \tsci{0.306}{[0.295, 0.316]} & \tsci{0.251}{[0.231, 0.270]} & \tsci{0.243}{[0.222, 0.264]}\\*
 &  & English & \tsci{0.326}{[0.316, 0.337]} & \tsci{0.195}{[0.178, 0.213]} & \tsci{0.173}{[0.155, 0.190]}\\*
 &  & Program & \tsci{0.277}{[0.268, 0.285]} & \tsci{0.279}{[0.259, 0.297]} & \tsci{0.452}{[0.421, 0.483]}\\
\midrule
10,000 & ViT & Japanese & \tsci{0.070}{[0.062, 0.077]} & \tsci{0.805}{[0.788, 0.822]} & \tsci{0.810}{[0.790, 0.827]}\\*
 &  & English & \tsci{0.142}{[0.133, 0.152]} & \tsci{0.515}{[0.493, 0.535]} & \tsci{0.485}{[0.462, 0.507]}\\*
 &  & Program & \tsci{0.047}{[0.042, 0.052]} & \tsci{0.876}{[0.862, 0.889]} & \tsci{0.889}{[0.874, 0.902]}\\
\addlinespace[0.25ex]
10,000 & VGG masks & Japanese & \tsci{0.215}{[0.205, 0.225]} & \tsci{0.393}{[0.371, 0.414]} & \tsci{0.354}{[0.329, 0.376]}\\*
 &  & English & \tsci{0.245}{[0.235, 0.255]} & \tsci{0.324}{[0.301, 0.342]} & \tsci{0.266}{[0.244, 0.285]}\\*
 &  & Program & \tsci{0.107}{[0.100, 0.115]} & \tsci{0.719}{[0.698, 0.738]} & \tsci{0.770}{[0.750, 0.790]}\\
\addlinespace[0.25ex]
10,000 & VGG no masks & Japanese & \tsci{0.199}{[0.189, 0.209]} & \tsci{0.477}{[0.454, 0.498]} & \tsci{0.463}{[0.438, 0.487]}\\*
 &  & English & \tsci{0.240}{[0.230, 0.250]} & \tsci{0.348}{[0.328, 0.370]} & \tsci{0.308}{[0.286, 0.330]}\\*
 &  & Program & \tsci{0.173}{[0.164, 0.182]} & \tsci{0.547}{[0.524, 0.570]} & \tsci{0.641}{[0.615, 0.667]}\\
\end{longtable}
\endgroup

\begingroup
\small
\begin{longtable}{@{}rllrrrrr@{}}
\caption{\label{tab:train_size_factors}Reduced-training-data metrics on the
standard test data, matching Table~\ref{tab:factor_results}.}\\
\toprule
Train & Model & Slice & N & Acc & C Acc & R Acc & M Acc\\
\midrule
\endfirsthead
\caption[]{Reduced-training-data metrics (continued).}\\
\toprule
Train & Model & Slice & N & Acc & C Acc & R Acc & M Acc\\
\midrule
\endhead
\midrule
\multicolumn{8}{r}{\small Continued on the next page.}\\
\endfoot
\bottomrule
\endlastfoot
2,500 & ViT & All & 2000 & 0.424 & 0.992 & 0.965 & 0.436\\*
 &  & Contained & 1431 & 0.416 & 0.992 & 0.955 & 0.431\\*
 &  & Containerless & 569 & 0.445 & \textsc{n/a} & 0.988 & 0.448\\
\addlinespace[0.25ex]
2,500 & VGG masks & All & 2000 & 0.276 & 0.997 & 0.950 & 0.280\\*
 &  & Contained & 1431 & 0.336 & 0.997 & 0.945 & 0.340\\*
 &  & Containerless & 569 & 0.123 & \textsc{n/a} & 0.963 & 0.128\\
\addlinespace[0.25ex]
2,500 & VGG no masks & All & 2000 & 0.146 & 0.990 & 0.935 & 0.154\\*
 &  & Contained & 1431 & 0.164 & 0.990 & 0.934 & 0.169\\*
 &  & Containerless & 569 & 0.102 & \textsc{n/a} & 0.938 & 0.116\\
\midrule
5,000 & ViT & All & 2000 & 0.691 & 0.999 & 0.986 & 0.696\\*
 &  & Contained & 1431 & 0.684 & 0.999 & 0.981 & 0.691\\*
 &  & Containerless & 569 & 0.707 & \textsc{n/a} & 0.998 & 0.707\\
\addlinespace[0.25ex]
5,000 & VGG masks & All & 2000 & 0.479 & 0.997 & 0.972 & 0.485\\*
 &  & Contained & 1431 & 0.564 & 0.997 & 0.966 & 0.571\\*
 &  & Containerless & 569 & 0.265 & \textsc{n/a} & 0.988 & 0.271\\
\addlinespace[0.25ex]
5,000 & VGG no masks & All & 2000 & 0.279 & 0.997 & 0.966 & 0.282\\*
 &  & Contained & 1431 & 0.316 & 0.997 & 0.960 & 0.317\\*
 &  & Containerless & 569 & 0.185 & \textsc{n/a} & 0.979 & 0.195\\
\midrule
10,000 & ViT & All & 2000 & 0.876 & 0.999 & 0.992 & 0.880\\*
 &  & Contained & 1431 & 0.868 & 0.999 & 0.990 & 0.874\\*
 &  & Containerless & 569 & 0.896 & \textsc{n/a} & 0.998 & 0.896\\
\addlinespace[0.25ex]
10,000 & VGG masks & All & 2000 & 0.719 & 1.000 & 0.986 & 0.722\\*
 &  & Contained & 1431 & 0.770 & 1.000 & 0.982 & 0.774\\*
 &  & Containerless & 569 & 0.589 & \textsc{n/a} & 0.996 & 0.591\\
\addlinespace[0.25ex]
10,000 & VGG no masks & All & 2000 & 0.547 & 0.998 & 0.978 & 0.552\\*
 &  & Contained & 1431 & 0.598 & 0.998 & 0.971 & 0.602\\*
 &  & Containerless & 569 & 0.420 & \textsc{n/a} & 0.993 & 0.427\\
\end{longtable}
\endgroup

\subsection{Initial VGG program-label failure}
\label{subapp:vgg_failure}

An initial masked VGG n-gram-2 program baseline, trained without label smoothing
or weight decay, exposed a failure mode that aggregate string metrics would
otherwise compress into a single low score. At the time, this motivated checking
similar VGG models that do not exhibit the same collapse.

Table~\ref{tab:vgg_decoder_ablation} compares this initial run with masked and
no-mask VGG variants and ViT. The initial masked n-gram-2 model learns the
program schema and recovers the container and modifier but obtains 0.000
contained accuracy and 0.000 contained motif accuracy. The closest tested masked
configuration that removed the collapse was the regularized n-gram-1 model
selected on composite validation error. Removing masks under the same
regularized composite-selection setup also removed the collapse: no-mask
n-gram-2 reached 0.737 contained motif accuracy. The no-mask baseline uses a
wider n-gram-4 context.

\begin{table}[H]
\centering
\caption{\label{tab:vgg_decoder_ablation}VGG n-gram-2 program-label failure and
  comparison with other models.}
\begin{tabular}{@{}lrrrr@{}}
\toprule
Model and setting & TER & Acc & Contained Acc & Contained M\\
\midrule
VGG, n-gram-2 masks & 0.317 & 0.151 & 0.000 & 0.000\\
VGG, n-gram-2 no masks + reg. & 0.106 & 0.722 & 0.731 & 0.737\\
VGG, n-gram-1 masks + reg. & 0.068 & 0.821 & 0.837 & 0.844\\
VGG, n-gram-4 no masks + reg. & 0.101 & 0.732 & 0.765 & 0.769\\
ViT & 0.022 & 0.941 & 0.932 & 0.938\\
\bottomrule
\end{tabular}
\end{table}

Table~\ref{tab:program_confusions} reports the motif prediction concentration
behind the initial VGG failure. The contained target motifs are highly
dispersed, but the initial n-gram-2 VGG program model maps the entire contained
slice to one motif token. This collapsed token, \texttt{M:0808},
appears only 13 times in the training split and never appears as a target
motif in the development or test splits.

\begin{table}[H]
\centering
\small
\caption{\label{tab:program_confusions}Motif prediction concentration in the
  initial n-gram-2 VGG collapsed program-label outputs.  `Top target' and
  `Top pred.' give the most common target and predicted motif tokens in each
  slice, with counts in parentheses.}
\begin{tabular}{@{}llrrlll@{}}
\toprule
Model & Slice & N & M Acc & Top target & Top pred. & Top wrong pred.\\
\midrule
Initial VGG n2 & All & 2000 & \textsc{n/a} & \texttt{M:00328} (5) & \texttt{M:0808} (1431) &
  \texttt{M:0808} (1431)\\
    & Contained & 1431 & 0.000 & \texttt{M:24407} (4) &
  \texttt{M:0808} (1431) & \texttt{M:0808} (1431)\\
    & Containerless & 569 & 0.529 & \texttt{M:26611} (3) &
  \texttt{M:1053} (4) & \texttt{M:1053} (4)\\
\midrule
ViT & All & 2000 & 0.946 & \texttt{M:00328} (5) & \texttt{M:00328} (5) &
  \texttt{M:26704} (3)\\
    & Contained & 1431 & 0.938 & \texttt{M:24407} (4) &
  \texttt{M:24407} (4) & \texttt{M:26704} (3)\\
    & Containerless & 569 & 0.965 & \texttt{M:26611} (3) &
  \texttt{M:26611} (3) & \texttt{M:0748} (2)\\
\bottomrule
\end{tabular}
\end{table}

\subsection{Learned masks for masked VGG baselines}
\label{subapp:masks}

\begin{figure}[H]
\centering
\small
\begin{minipage}[t]{0.48\textwidth}
\centering
Japanese, 12 positions\\[0.25ex]
\includegraphics[width=\linewidth]{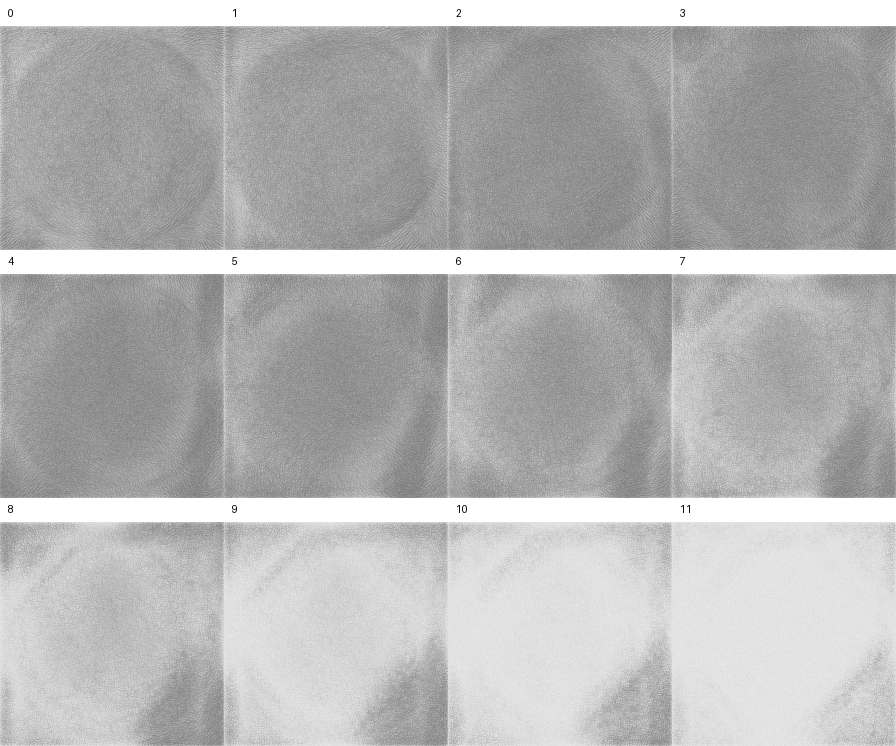}
\end{minipage}\hfill
\begin{minipage}[t]{0.48\textwidth}
\centering
English, 15 positions\\[0.25ex]
\includegraphics[width=\linewidth]{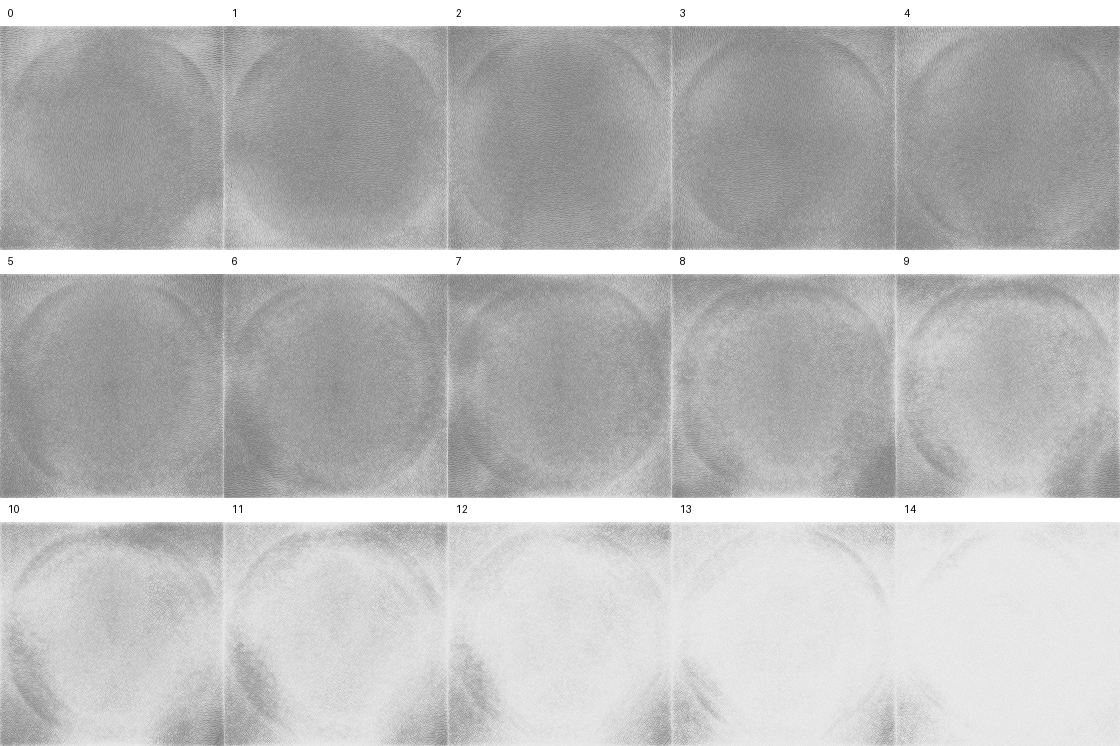}

\vspace{0.8ex}
Program, 4 positions\\[0.25ex]
\includegraphics[width=\linewidth]{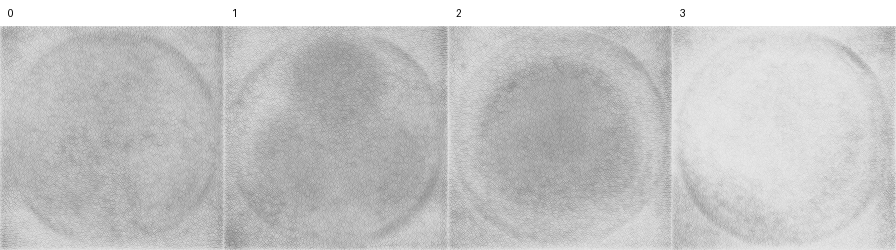}
\end{minipage}
\caption{Positional masks from the masked VGG baselines. Images are inverted:
  darker regions indicate larger retained mask values.}
\label{fig:masks}
\end{figure}

Figure~\ref{fig:masks} shows learned position-dependent masks for
the masked VGG baselines.

\clearpage

\subsection{LLM prompt for 20 test examples}
\label{subapp:llm_prompt_20ex}

The following prompt, reformatted for display, contains five labeled kamon
demonstrations followed by 20 randomly sampled synthetic test queries.
The resulting outputs are shown in Table~\ref{tab:llms}.

\noindent\makebox[\textwidth]{\begin{minipage}{380bp}
\documentimage{more_test_examples}{1}{60bp 474.8898bp 60.2756bp 60bp}
\par\vspace{6bp}\noindent
\begin{minipage}[t]{95bp}\centering\vspace{0pt}
\documentimage{more_test_examples}{1}{60bp 270.2398bp 470.2756bp 375bp}
\end{minipage}%
\begin{minipage}[t]{95bp}\centering\vspace{0pt}
\documentimage{more_test_examples}{1}{60bp 72.8898bp 470.2756bp 571.65bp}
\end{minipage}%
\begin{minipage}[t]{95bp}\centering\vspace{0pt}
\documentimage{more_test_examples}{2}{60bp 590.9398bp 470.2756bp 54bp}
\end{minipage}%
\begin{minipage}[t]{95bp}\centering\vspace{0pt}
\documentimage{more_test_examples}{2}{60bp 393.8898bp 470.2756bp 250.95bp}
\end{minipage}
\par\vspace{6bp}\noindent
\documentimage{more_test_examples}{2}{60bp 285.8898bp 86.2756bp 455bp}
\end{minipage}}\par

\clearpage

\subsection{Few-shot multimodal LLM performance}
\label{subsec:llm_app}

Table~\ref{tab:llms} shows the Japanese descriptions produced by masked VGG, ViT,
Claude~Opus~4.7~Max and GPT~5.4~xhigh for the 20 synthetic test queries in
Supplementary Materials~\ref{subapp:llm_prompt_20ex}. Check marks indicate
exact matches to the target description; other cells show the generated text.

\begin{table}[H]
\centering
\caption{\label{tab:llms}Japanese descriptions generated for 20 synthetic test
  crests. Masked VGG and ViT are supervised baselines; Claude and GPT receive five
  labeled kamon demonstrations without fine-tuning on KamonBench. Check marks
  indicate exact matches to the target description.}
~\\
{\scriptsize
\setlength{\tabcolsep}{3pt}
\begin{tabular}{@{}lp{0.32\linewidth}p{0.11\linewidth}cp{0.20\linewidth}p{0.20\linewidth}@{}}
\toprule
Image & Description/Translation & VGG masks & ViT & Claude & GPT\\
\midrule
\includegraphics[width=0.5cm]{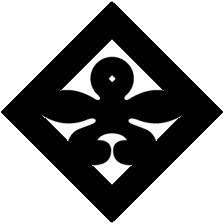} & 隅立て角に秋津洲浜/Akitsushima hama in a corner-standing corner &
$\checkmark$ & $\checkmark$ & 隅立て井桁に花菱 & 唐花菱\\
\includegraphics[width=0.5cm]{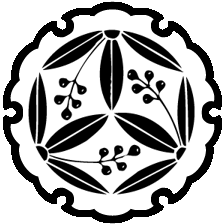} & 雪輪に三つ割り南天/three-split nanten in a snow ring &
$\checkmark$ & $\checkmark$ & 雪輪に頭合せ三つ茗荷 & 木瓜に三つ南天\\
\includegraphics[width=0.5cm]{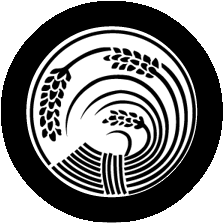} & 丸に左稲の丸/circle of left rice plant in a circle &
$\checkmark$ & $\checkmark$ & 丸に稲 & 丸に抱き稲\\
\includegraphics[width=0.5cm]{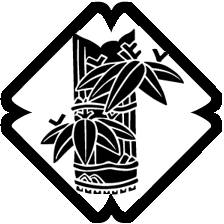} & 細隅入り角に変わり切り竹笹/modified cut bamboo grass in a thin corner-entering square &
$\checkmark$ & $\checkmark$ & 雪輪菱に笹 & 竹菱\\
\includegraphics[width=0.5cm]{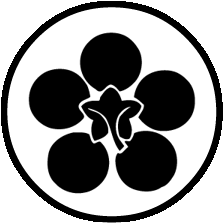} & 糸輪に台梅鉢/stand plum blossom crest in a thread ring &
$\checkmark$ & $\checkmark$ & 丸に梅鉢 & 丸に五瓜に唐花\\
\includegraphics[width=0.5cm]{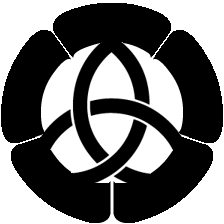} & 瓜輪に結び柏/tied-together oak leaves in a melon ring &
$\checkmark$ & $\checkmark$ & 丸に三つ輪違い & 三つ葵巴\\
\includegraphics[width=0.5cm]{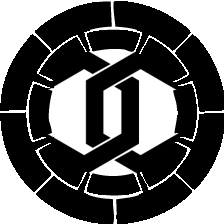} & 源氏輪に違い一重亀甲/alternating single-layered tortoiseshell pattern in a Genji ring &
$\checkmark$ & $\checkmark$ & 丸に輪宝 & 源氏車に輪違い\\
\includegraphics[width=0.5cm]{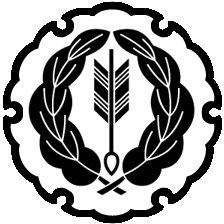} & 雪輪に江原柏/Ebara oak leaves in a snow ring &
$\checkmark$ & $\checkmark$ & 雪輪に矢 & 木瓜に抱き矢\\
\includegraphics[width=0.5cm]{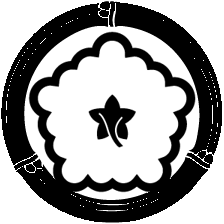} & 竹輪に中陰裏唐花/medium-shaded reverse karakusa in a bamboo ring &
$\checkmark$ & $\checkmark$ & 雪輪に花菱 & 丸に木瓜に唐花\\
\includegraphics[width=0.5cm]{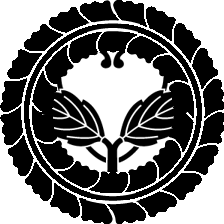} & 藤輪に二枚葉蔓梶の葉/two-leaves vine kaji leaves in a wisteria ring &
$\checkmark$ & $\checkmark$ & 丸に抱き茗荷 & 丸に抱き柏\\
\includegraphics[width=0.5cm]{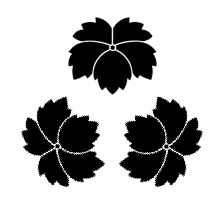} & 尻合せ三つ大割鬼蔦/bottom-joined three large split demon ivy &
$\checkmark$ & $\checkmark$ & 三つ盛り柏 & 尻合せ三つ葵\\
\includegraphics[width=0.5cm]{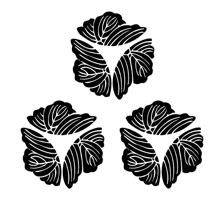} & 三つ盛り外三つ割り蔦/three-piled outer-three-split ivy &
$\checkmark$ & $\checkmark$ & 頭合せ三つ柏 & 尻合せ三つ蔦\\
\includegraphics[width=0.5cm]{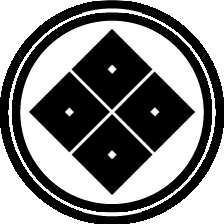} & 二重輪に隅立て四つ目/corner-standing four-eyed in a double ring &
$\checkmark$ & $\checkmark$ & 丸に武田菱 & 丸に四つ目菱\\
\includegraphics[width=0.5cm]{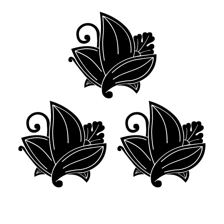} & 三つ盛り南天蝶/three piled nanten butterflies &
$\checkmark$ & $\checkmark$ & 三つ盛り蔦 & 三つ盛り茗荷\\
\includegraphics[width=0.5cm]{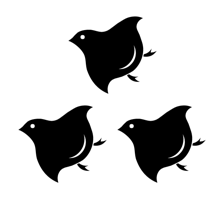} & 三つ盛り千鳥/three piled chidori &
尻合せ三つ千鳥 & $\checkmark$ & 三つ盛り雀 & $\checkmark$\\
\includegraphics[width=0.5cm]{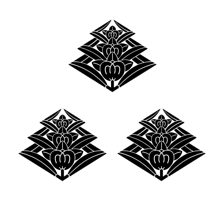} & 三つ盛り三階菱橘/three-piled three-tiered water caltrop tachibana &
$\checkmark$ & $\checkmark$ & 三つ盛り割菱 & 三つ盛り三階菱\\
\includegraphics[width=0.5cm]{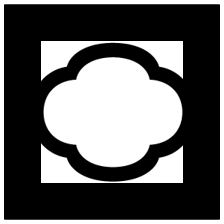} & 平角に木瓜形/quince shape in a flat square &
$\checkmark$ & $\checkmark$ & 石持ち地抜き木瓜 & 角に木瓜\\
\includegraphics[width=0.5cm]{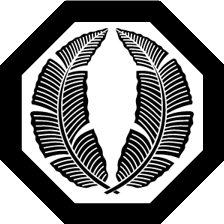} & 八角に抱き芭蕉/embracing banana plant in an octagon &
$\checkmark$ & $\checkmark$ & 八角に抱き笹 & 八角に違い鷹の羽\\
\includegraphics[width=0.5cm]{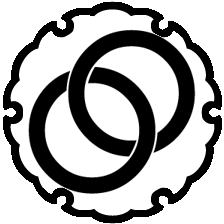} & 雪輪に寄り懸り輪違い/alternating leaning-hanging rings in a snow ring &
$\checkmark$ & $\checkmark$ & 雪輪に輪違い & 木瓜に輪違い\\
\includegraphics[width=0.5cm]{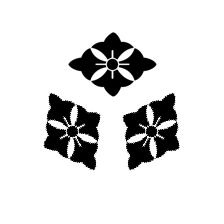} & 頭合せ三つ利休花菱/head-joined three Rikyū flower diamonds &
$\checkmark$ & $\checkmark$ & 三つ盛り花菱 & 尻合せ三つ桔梗\\
\bottomrule
\end{tabular}
} 
\end{table}

\clearpage

\subsection{Human instructions: English}
\label{subapp:human_instr_en}

\appendiximage{instructions_en}{1}{73.8898}{60}
\noindent\begin{minipage}{\textwidth}
\appendiximage{instructions_en}{2}{135.8898}{54}
\vspace{6bp}
\appendiximage{instructions_en}{3}{646.8898}{54}
\end{minipage}\par

\subsection{Human instructions: Japanese}
\label{subapp:human_instr_ja}

\appendiximage{instructions_ja}{1}{57.8898}{60}
\noindent\begin{minipage}{\textwidth}
\appendiximage{instructions_ja}{2}{82.8898}{54}
\vspace{6bp}
\appendiximage{instructions_ja}{3}{746.8898}{54}
\end{minipage}\par

\subsection{Questionnaire sample}
\label{subapp:human_quest}

\begin{center}
\includegraphics[width=0.75\textwidth]{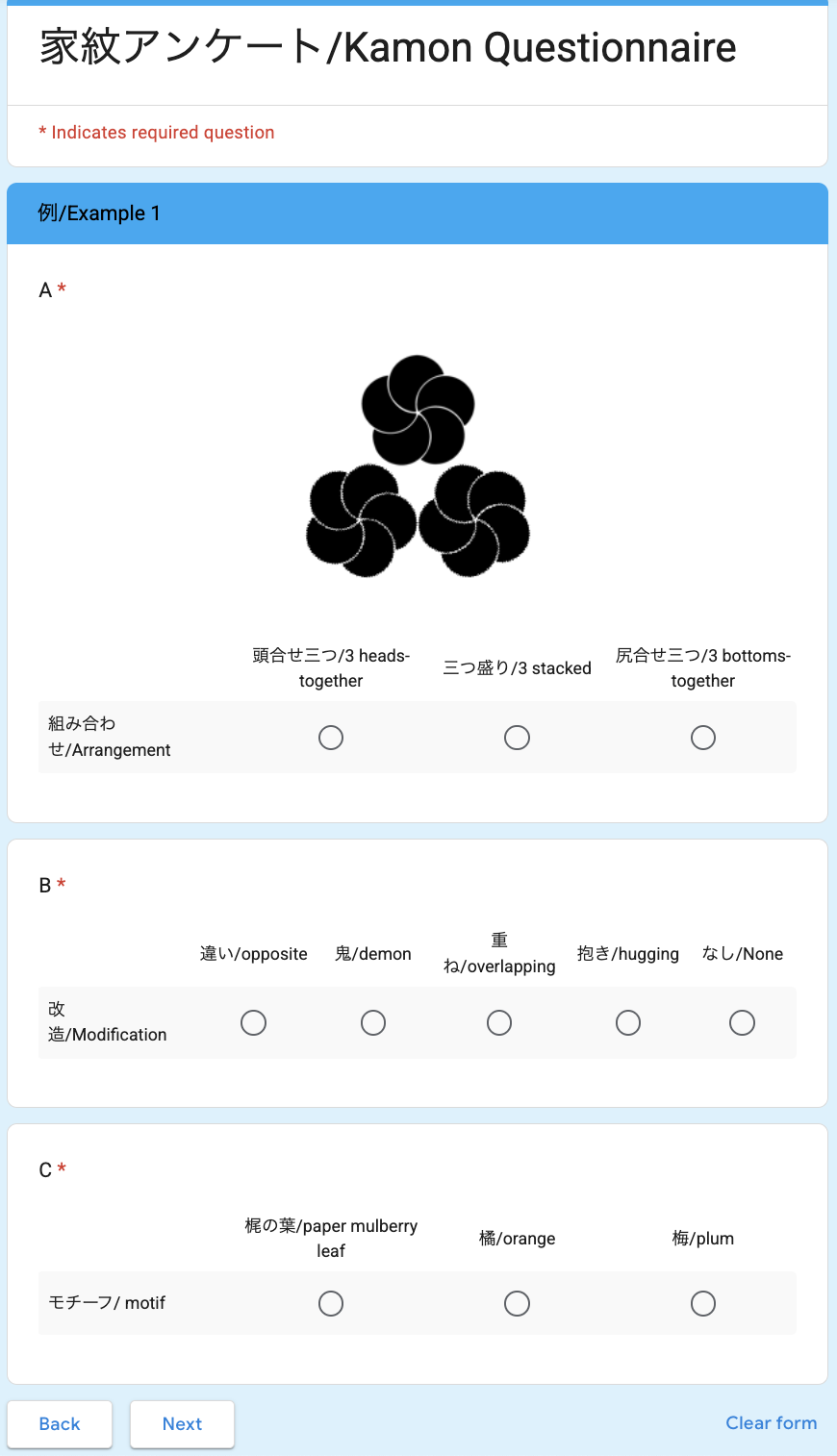}
\end{center}

\clearpage
\subsection{LLM prompt for 10 kamon rated by humans}
\label{subapp:llm_prompt_10}

The prompt below is reformatted for display.

\appendiximage{test_examples}{1}{71.8898}{60}
\noindent\begin{minipage}{\textwidth}
\appendiximage{test_examples}{2}{70.8898}{54}
\vspace{6bp}
\appendiximage{test_examples}{3}{706.8898}{54}
\end{minipage}\par

\noindent\makebox[\textwidth]{\begin{minipage}{380bp}
\documentimage{test_examples}{3}{60bp 644.8898bp 60.2756bp 142bp}
\par\vspace{6bp}\noindent
\begin{minipage}[t]{190bp}\centering\vspace{0pt}
\documentimage{test_examples}{3}{60bp 63.8898bp 395.2756bp 204bp}
\end{minipage}%
\begin{minipage}[t]{190bp}\centering\vspace{0pt}
\documentimage{test_examples}{4}{60bp 212.8898bp 395.2756bp 54bp}
\end{minipage}
\par\vspace{6bp}\noindent
\documentimage{test_examples}{4}{60bp 104.8898bp 86.2756bp 635bp}
\end{minipage}}\par


\end{document}